\documentclass[10pt,twocolumn,letterpaper]{article}

\usepackage{cvpr}   
\usepackage{amsmath,amssymb,mathtools,booktabs,multirow,graphicx,multirow,amssymb,caption,pifont}
\usepackage{algorithm,algpseudocode,subcaption}
\usepackage[accsupp]{axessibility}

\definecolor{cvprblue}{rgb}{0.21,0.49,0.74}
\usepackage[pagebackref,breaklinks,colorlinks,allcolors=cvprblue]{hyperref}

\def\paperID{14402} 
\def\confName{CVPR}
\def\confYear{2026}

\title{Momentum Memory for Knowledge Distillation in Computational Pathology}

\author{
    Yongxin Guo$^{1}$\thanks{Corresponding author: Yongxin.Guo@wfusm.edu.\\ Code: \url{https://github.com/CAIR-LAB-WFUSM/MoMKD}}, \ 
    Hao Lu$^{1}$, \ 
    Onur C. Koyun$^{1}$, \ 
    Zhengjie Zhu$^{1}$, \ 
    Muhammet F. Demir$^{1}$, \ 
    Metin N. Gurcan$^{1}$ \\[4pt]
    $^{1}$Wake Forest University School of Medicine, Winston-Salem, NC, USA \\[4pt]
    {\tt\small \{Hao.Lu, Onur.Koyun, muhammet.demir\}@advocatehealth.org} \\
    {\tt\small \{Yongxin.Guo, Zhengjie.Zhu, Metin.Gurcan\}@wfusm.edu}
}

\begin{document}
\maketitle

\begin{abstract}
Multimodal learning that integrates genomics and histopathology has shown strong potential in cancer diagnosis, yet its clinical translation is hindered by the limited availability of paired histology–genomics data. Knowledge distillation (KD) offers a practical solution by transferring genomic supervision into histopathology models, enabling accurate inference using histology alone. However, existing KD methods rely on batch-local alignment, which introduces instability due to limited within-batch comparisons and ultimately degrades performance.
To address these limitations, we propose Momentum Memory Knowledge Distillation (MoMKD), a cross-modal distillation framework driven by a momentum-updated memory. This memory aggregates genomic and histopathology information across batches, effectively enlarging the supervisory context available to each mini-batch. Furthermore, we decouple the gradients of the genomics and histology branches, preventing genomic signals from dominating histology feature learning during training and eliminating the modality-gap issue at inference time.
Extensive experiments on the TCGA-BRCA benchmark (HER2, PR, and ODX classification tasks) and an independent in-house testing dataset demonstrate that MoMKD consistently outperforms state-of-the-art MIL and multimodal KD baselines, delivering strong performance and generalization under histology-only inference. Overall, MoMKD establishes a robust and generalizable knowledge distillation paradigm for computational pathology. 
\end{abstract}

\section{Introduction}
\label{sec:intro}

Deep multiple instance learning (MIL) models trained solely on histopathology slides have achieved strong performance~\cite{gurcanHistopathologicalImageAnalysis2009,suComputationalPathologyAccurate2024,zhuDGRMILExploringDiverse2025,guoBPMambaMILBioinspiredPrototypeguided2025}, yet their predictive ceiling remains constrained by the limited information in the image representation. Many clinically important biomarkers, for example, molecular subtypes, gene-expression signatures, or recurrence risk scores, are defined not by visual appearance but by underlying molecular signals \cite{yuMultimodalDataFusion2024}. High-throughput transcriptomic assays can capture this information directly \cite{baysoyTechnologicalLandscapeApplications2023,nemaOmicsbasedTumorMicroenvironment2024}. Multimodal learning that integrates genomics with histopathology has shown improved performance by combining these complementary sources of information~\cite{chen2021multimodal,chen2022pan,Xu_2023_ICCV, gong2025medcmr}. However, genomic data are expensive and slow, especially in resource-limited settings. To circumvent this limitation, recent studies have explored knowledge distillation (KD) to inject genomic information into histology models during training, enabling histology-only inference while retaining molecular predictive power.

\begin{figure}[t]
    \centering
    
    \includegraphics[width=\linewidth, keepaspectratio,
    , trim=180 80 190 40, clip]{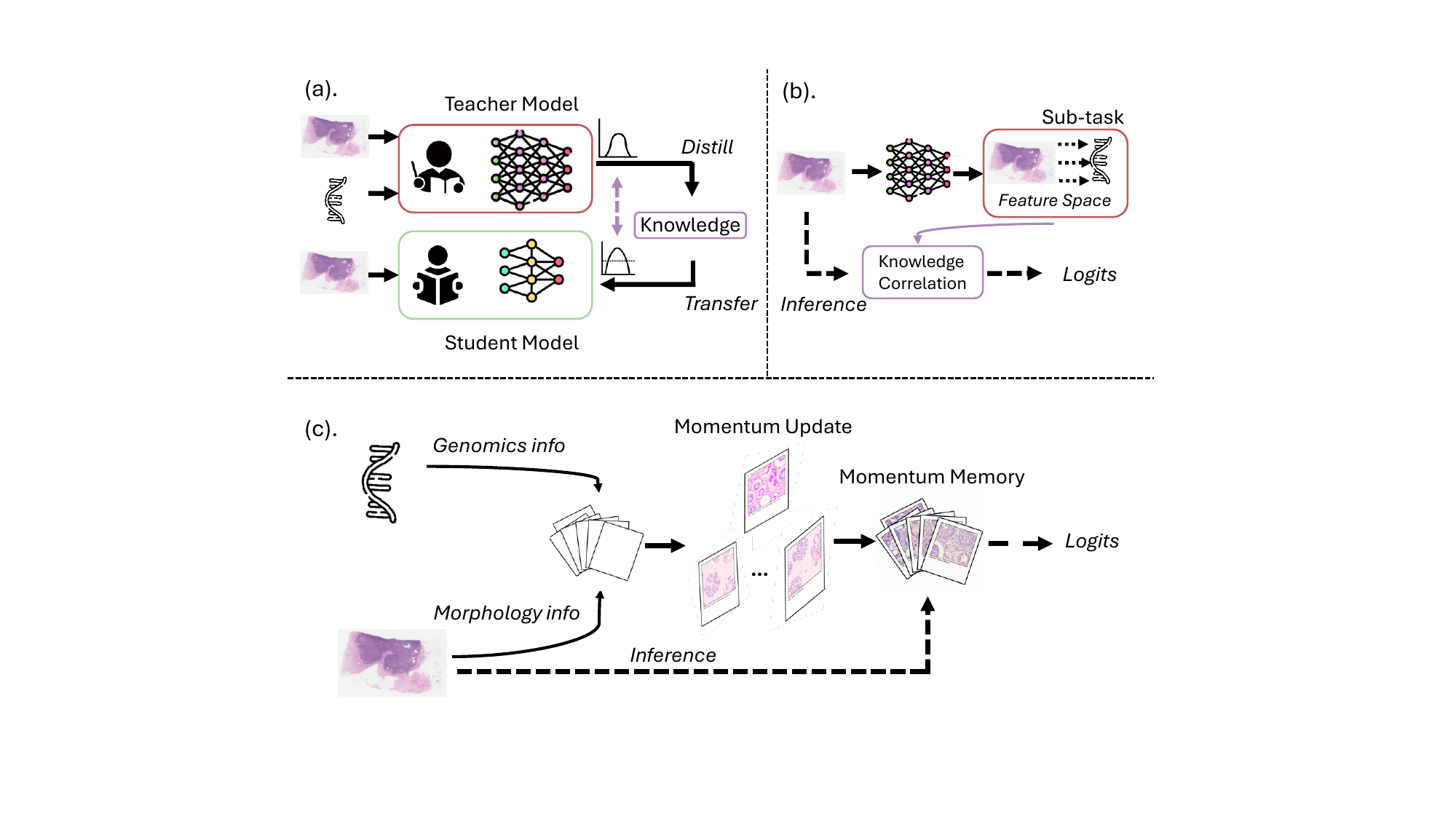}
    \caption{a. The classical teacher-student knowledge distillation (KD) method. b. The correlation-based KD method. c. The proposed momentum memory KD method. Compared with a and b, the proposed method uses momentum memory for knowledge distillation to solve the batch-local problem.}
    \label{fig:overview}
\end{figure}

Existing KD frameworks transfer supervision from a multimodal teacher (e.g., histology with omics) to a unimodal student (histology-only)~\cite{hintonDistillingKnowledgeNeural2015,maGeneralizablePathologyFoundation2025,xingComprehensiveLearningAdaptive2024,chenLearningPrivilegedMultimodal2022}. However, most existing frameworks in pathology domain operationalize KD as \emph{batch-local, continuous feature matching}: either forcing heterogeneous modalities map into the same latent space~\cite{xingComprehensiveLearningAdaptive2024, maGeneralizablePathologyFoundation2025,10530449} (Fig.~\ref{fig:overview}a) or regressing auxiliary cross-modal targets and treating them as distilled knowledge~\cite{wangHistoGenomicKnowledgeDistillation2024,li2025unleashing} (Fig.~\ref{fig:overview}b). Such intra-batch supervision is inherently fragile. Its supervision signal is transient, which is defined only by the current mini-batch, and offers limited negative sample diversity. This forces an unstable, direct alignment between asymmetric modalities (e.g., genomics data and pathology image). This fragility is further amplified in the pathological MIL setting. The information redundancy of gigapixel images, where noisy background regions dominate the mini-batch, simply overwhelms the distillation signal, leading to brittle generalization under domain shift.

To address this limitation, we introduce a momentum-updated memory bank that accumulates cross-modal representations over the entire training trajectory, providing each mini-batch with a much richer supervisory context. We propose Momentum Memory for Knowledge Distillation (MoMKD), which maintains a slowly evolving, label-conditioned momentum memory. Similarly to dynamic dictionaries in contrastive learning~\cite{heMomentumContrastUnsupervised2020}, the memory aggregates genomics-histopathology information across the entire training trajectory. It simultaneously serves as an information bottleneck that compresses redundant histology features and as an distillation mediator that injects genomic semantics into histology representations (Fig.~\ref{fig:overview}c). MoMKD’s operates through two key mechanisms. First, a soft angle-based loss aligns both modalities to the shared memory space, enlarging the supervisory context across each mini-batch. Second, a gradient-decoupled update strategy isolates the genomic and histology branches, preventing genomic gradients from overwhelming histology features during training and eliminating the modality gap in the uni-modal inference stage.

We validate MoMKD across three classification tasks (HER2, PR, and Oncotype DX) on the TCGA-BRCA benchmark and an independent institutional cohort. MoMKD consistently outperforms state-of-the-art MIL and multimodal KD baselines, demonstrating superior cross-domain generalization. Overall, MoMKD introduces a new cross-modal distillation paradigm that replaces high-variance batch-level alignment with a stable, compact momentum memory, enabling robust multimodal knowledge distillation.

In summary, our contributions are threefold:
\begin{enumerate}
    \item \textbf{Momentum memory for cross-modal distillation:} A label-conditioned, dynamically evolving dictionary that accumulates genomics-histopathology statistics, replacing stochastic batch-local matching with stable dictionary-based alignment.
  \item \textbf{Gradient-decoupled optimization:} A decoupling strategy that isolates genomics and histology gradients, preventing modality-gap issues between multimodal training and unimodal inference.
  \item \textbf{Extensive validation and analysis:} Demonstrating strong performance and generalization across datasets, with visualizations showing that the learned memory captures meaningful biological structures.
\end{enumerate}
\section{Related Works}
\label{sec:related}

\subsection{Application of MIL in WSIs}
Multiple Instance Learning (MIL) is the prevailing paradigm in computational pathology: a WSI is treated as a bag of instances with slide-level supervision, enabling weakly supervised prediction. Early and influential methods like ABMIL \cite{ilseAttentionbasedDeepMultiple2018} introduced attention mechanisms to learn the importance of each instance for the slide-level prediction. Subsequent work such as CLAM \cite{luDataefficientWeaklySupervised2021a}, improves the attention mechanism for better interpretability, and transformer-based architectures leveraging self-attention to model long-range dependencies between patches \cite{shaoTransMILTransformerBased2021,biMILViTMultipleInstance2023}.

To better capture the tumor microenvironment, recent approaches have shifted towards graph-based MIL \cite{liDynamicGraphRepresentation2024,ahmedt-aristizabalSurveyGraphbasedDeep2022}. This has led to improved performance on tasks that rely on contextual information. However, a fundamental limitation persists across all WSI-only frameworks: their performance is bounded by the morphological information present in the H\&E stain. For tasks inherently defined by molecular state (\eg, gene expression-based risk scores), the visual signal is often insufficient, creating a performance ceiling that motivates integrating privileged molecular data.

\begin{figure*}[t]
    \centering
    \includegraphics[width=\textwidth, trim=70 15 105 5, clip]{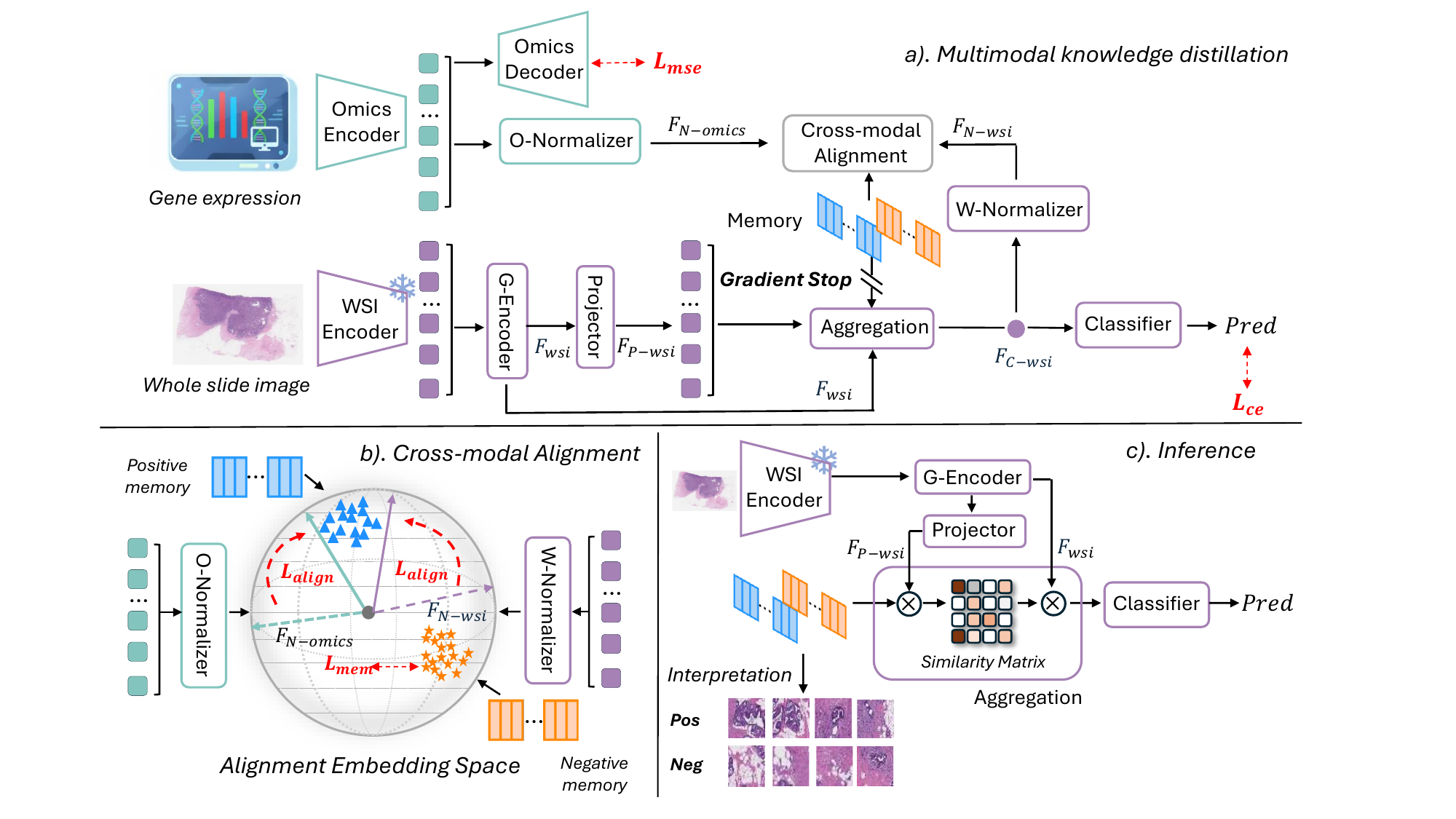}
    
    \caption{The overall framework of the proposed momentum memory knowledge distillation framework. a) presents the multi-modal KD training process; b) indicates the proposed cross-modal alignment. Here we assume the input is a positive case so that the alignment loss is aiming to push it closer to the positive memory set (blue triangle) and pull it away to the negative memory set (yellow star); c) presents the uni-modal inference stage, and the visual interpretation on the memory.}
    \label{fig:framework} 
\end{figure*}

\subsection{Multi-modal Knowledge Distillation for WSI analysis}
To break the performance ceiling of WSI-only models, the Knowledge Distillation (KD) which leverages paired histo-genomic data during training while keeps inference WSI-only has emerged as a pragmatic solution~\cite{gindraLargeScaleBenchmarkCrossModal2025, zhangDisentangledMultimodalLearning2025}. In the domain of pathology image analysis, teacher-student method is one of the most popular streams~\cite{bontempoGraphBasedMultiScaleApproach2024,wangDistillingHeterogeneousKnowledge2025,shuSlideGCDSlideBasedGraph2024,lan2026recokd,lan2025acamkd}. For example, Xing et al.~\cite{xingComprehensiveLearningAdaptive2024} proposed a gradient-based teacher-student framework for the tumor grading task. Compared with this two-stage KD method, Zhang et al. \cite{zhangMultimodalKnowledgeDecomposition2025} proposed an online learning paradigm based on a multi-teacher framework for biomarker prediction. Additionally, G-HANet \cite{wangHistoGenomicKnowledgeDistillation2024} reconstructed functional genomic information from WSI via cross-modal learning and conducted distillation for survival analysis.

Despite progress, batch-local feature regression limits cross-modal KD: supervision hinges on mini-batch idiosyncrasies, limited-class samples are confined within a single batch, and heterogeneous geometries are forced to match, yielding brittle generalization under shift. This exact challenge was a critical hurdle in self-supervised representation learning. The field's evolution provides a clear blueprint: SimCLR demonstrated the need for massive batches~\cite{chenSimpleFrameworkContrastive2020,li2025pointdico,sun2025hyperpoint,su2025medgrpomultitaskreinforcementlearning}, while memory banks~\cite{wuUnsupervisedFeatureLearning2018} and dynamic dictionaries, such as MoCo~\cite{heMomentumContrastUnsupervised2020} and lots of variations~\cite{zhangUSERUnifiedSemantic2024,heMoCoSAMomentumContrast2024,2f3f68f41271474c8e5638b13b6b8f8f} showed that a large and consistent dictionary is the key to stable learning and robust transferability. Inspired by this, we introduce MoMKD, which adapts the memory look-up concept to the cross-modal KD challenge. The momentum memory acts as a dynamic, global dictionary, accumulating genomics-histopathology statistics over the entire training trajectory which allows MoMKD to align modalities via a stable, shared semantic space, rather than through direct and noisy feature matching.

\section{Methodology}
\label{sec:methodology}
\subsection{Overview}

Computational pathology faces an intrinsic asymmetry in cross-modal distillation: global transcriptomic embeddings versus local histopathology representations. Direct and batch-local feature alignment magnifies modality noise and leads to unstable optimization.

To overcome this, we reformulate cross-modal alignment by learning a momentum memory as the distillation mediator rather than forcing direct feature matching. Specifically, both modalities interact through a compact and information-rich memory  which capturing canonical genomics–histopathology joint concepts (Fig.~\ref{fig:framework}). In the following section, we will introduce more details of the proposed method.

\subsection{Dual-Branch Encoding}
\label{sec:dual_branch}
MoMKD employs a graph-based WSI encoder to model the tumor microenvironment~\cite{liDynamicGraphRepresentation2024,chen2021whole}, integrated with a fully connected omics encoder. Both modalities are projected into a shared, $L_2$-normalized latent space for cross-modal alignment while preserving modality-specific structure.

\paragraph{Graph-based WSI encoder.}
For a given WSI $x_i$, we use a frozen encoder (e.g., UNI v2~\cite{chenGeneralpurposeFoundationModel2024}) to obtain patch embeddings. Then, each WSI is represented as a spatial graph $G=(V,E)$, where nodes $V=\{v_1,\dots,v_I\}$ correspond to image patches and edges $E$ connect each node to its $k$-nearest neighbors ($k{=}8$) based on centroid distance. 
A two-layer GATv2 (Refered to the G-Encoder in Fig.~\ref{fig:framework}a)~\cite{brodyHowAttentiveAre2022} encodes contextualized patch embeddings
$F_{\mathrm{wsi}} = f_{\mathrm{wsi}}(x_i)\!\in\!\mathbb{R}^{I\times D}$ ($D{=}256$).
The projected features
$F_{\mathrm{P\text{-}wsi}} = W_{\mathrm{proj}} \cdot F_{\mathrm{wsi}}\in\mathbb{R}^{I\times D}$
are compared with the memory $\mathcal{C}$ to produce similarity scores.
Aggregating these patch-level similarities yields a slide-level representation $F_{\mathrm{C\text{-}wsi}}$.
For the cross-modal alignment, $F_{\mathrm{C\text{-}wsi}}$ will be linearly projected onto a spherical space as the WSI representation:
\begin{equation}
\mathbf{F}_{\mathrm{N\text{-}wsi}} =
\frac{W_{\mathrm{C\text{-}wsi}} \cdot \mathbf{F}_{\mathrm{C\text{-}wsi}}}{\left\|W_{\mathrm{C\text{-}wsi}} \cdot \mathbf{F}_{\mathrm{C\text{-}wsi}}\right\|_2}
\in\mathbb{R}^{D_N}
\label{eq:f_n_wsi}
\end{equation}
where $W_{\mathrm{C\text{-}wsi}}$ denotes learnable weights, $D_N$=128.

\paragraph{Omics encoder.}
A lightweight MLP first projects the omics vector into the same latent space: $F_{\mathrm{omics}} = f_{\mathrm{omics}}(\text{omics}) \in \mathbb{R}^{D}$.
The $F_{\mathrm{omics}}$ is then transformed and $L_2$-normalized with a learnable matrix $W_{\mathrm{omics}}$:
\begin{equation}
\mathbf{F}_{\mathrm{N\text{-}omics}} =
\frac{W_{\mathrm{omics}} \cdot \mathbf{F}_{\mathrm{omics}}}{\left\|W_{\mathrm{omics}} \cdot \mathbf{F}_{\mathrm{omics}}\right\|_2}
\in\mathbb{R}^{D_N}
\label{eq:f_n_omics}
\end{equation}

\subsection{Momentum Knowledge-distillation via Cross-Modal Alignment}

\subsubsection{Momentum Memory as Knowledge Mediator}
A fundamental challenge in multi-modal learning is the modality gap. To solve this problem, we introduced a momentum memory. This compact set of class-conditional memory setting: $C^+$ and $C^-$ represent positive class and negative class with $n$ memory components ($c_j^+$ and $c_j^-$, $j=[1,...,n]$) serve as the distillation mediator and anchor both modalities to a common decision geometry. Within the training progress, more information from both omics and WSI will be compressed and accumulated to this memory, which represents the global semantic representation rather than a simple instance cache. Consequently, the model aligns heterogeneous features with this highly-compressed, slowly-evolving mediator, rather than chasing a noisy, rapidly-updating intra-batch distribution.

\begin{table*}[t]
    \centering
    \caption{Internal comparison on the TCGA-BRCA dataset. The best performance is in bold, and the second-best performance is underlined.}
    \label{tab:internal_results}
    \resizebox{\textwidth}{!}{
    \begin{tabular}{l ccc ccc ccc}
        \toprule
        \multirow{2}{*}{\textbf{Methods}} & \multicolumn{3}{c}{\textbf{HER2(\%)}} & \multicolumn{3}{c}{\textbf{PR(\%)}} & \multicolumn{3}{c}{\textbf{ODX(\%)}} \\
        \cmidrule(lr){2-4} \cmidrule(lr){5-7} \cmidrule(lr){8-10}
        & AUC & ACC & F1 & AUC & ACC & F1 & AUC & ACC & F1 \\
        \midrule
        ABMIL~\cite{ilseAttentionbasedDeepMultiple2018}   & 72.9$\pm$3.1 & 77.1$\pm$1.8 & 64.8$\pm$2.9 & 84.5$\pm$2.3 & 78.8$\pm$1.6 & 75.0$\pm$2.1 & 79.3$\pm$2.5 & 83.8$\pm$2.6 & 68.8$\pm$1.6 \\
        DSMIL~\cite{liDualstreamMultipleInstance2021}   & 71.3$\pm$4.3 & 76.2$\pm$2.8 & \underline{63.5$\pm$4.1} & 81.6$\pm$2.8 & 76.4$\pm$2.0 & 72.7$\pm$2.2 & 78.7$\pm$2.0 & 83.9$\pm$1.9 & 70.8$\pm$1.4 \\
        TransMIL~\cite{shaoTransMILTransformerBased2021}& 69.8$\pm$2.8 & 75.6$\pm$1.9 & 61.1$\pm$5.2 & 85.2$\pm$1.0 & 79.0$\pm$1.2 & 76.1$\pm$1.6 & 77.9$\pm$3.5 & 83.5$\pm$3.7 & 66.8$\pm$5.0 \\
        DTFDMIL~\cite{zhangDTFDMILDoubleTierFeature2022} & 74.4$\pm$1.9 & 76.2$\pm$1.0 & 61.7$\pm$5.4 & 83.9$\pm$2.4 & 79.6$\pm$3.1 & \underline{76.3$\pm$2.9} & 79.1$\pm$2.1 & 82.2$\pm$3.1 & 68.9$\pm$2.0 \\
        WIKG~\cite{liDynamicGraphRepresentation2024}    & 75.5$\pm$5.0 & 77.1$\pm$3.1 & 53.7$\pm$4.4 & 84.9$\pm$3.0 & 79.2$\pm$3.3 & 76.1$\pm$4.1 & 78.3$\pm$3.7 & \underline{84.1$\pm$1.2} & \underline{71.1$\pm$2.5} \\
        \midrule
        TDC~\cite{xingComprehensiveLearningAdaptive2024}     & 76.2$\pm$2.1 & 76.7$\pm$3.9 & 63.3$\pm$1.1 & 84.7$\pm$5.3 & 73.1$\pm$2.1 & 72.3$\pm$3.1 &\underline{ 81.0$\pm$2.2} & 81.6$\pm$3.3 & 70.9$\pm$2.4 \\
        MKD~\cite{zhangMultimodalKnowledgeDecomposition2025}     & \underline{77.1$\pm$2.3} & \underline{77.2$\pm$5.3} & 59.9$\pm$1.2 & \underline{85.1$\pm$1.2} & \underline{80.1$\pm$1.1} & 76.2$\pm$2.3 & 80.1$\pm$1.5 & 80.0$\pm$4.2 & 70.8$\pm$3.9 \\
        G-HANET~\cite{wangHistoGenomicKnowledgeDistillation2024} & 76.1$\pm$5.6 & 71.2$\pm$6.8 & 62.4$\pm$4.0 & 85.0$\pm$2.3 & 79.1$\pm$3.1 & 75.9$\pm$5.3 & 80.5$\pm$1.3 & 81.5$\pm$4.2 & 71.0$\pm$5.9 \\
        \midrule
        \textbf{MoMKD} & \textbf{79.6$\pm$0.7} & \textbf{77.9$\pm$4.5} & \textbf{67.8$\pm$3.3} & \textbf{87.9$\pm$0.9} & \textbf{81.0$\pm$2.3} & \textbf{78.8$\pm$1.9} & \textbf{82.3$\pm$2.3} & \textbf{85.6$\pm$0.8} & \textbf{74.9$\pm$1.9} \\
        \bottomrule
    \end{tabular}
    }
\end{table*}

\subsubsection{Indirect Memory-based Distillation}

Our distillation is an indirect process: omics and WSI features do not directly teach each other. Instead, both modalities are forced to align with the shared momentum memory. The overall workflow can be explained as follows:

Before training, we let the memory have a quick look at the image data: we randomly sample 10000 patches and use K-means clustering for initialization. By doing this, the starting epoch can be more stable and meaningful.

\paragraph{Semantic Anchoring via Omics Alignment.}
This step injects the visually initialized memory with genomics-grounded semantics. To ensure the omics representation itself is biologically faithful during training, we introduced a self-supervised reconstruction task to constrain the omics encoder. With a robust omics embedding thus guaranteed, the omics projection $F_{\text{N-omics}}$ is pushed to align with its corresponding memory (e.g., $C^+$) and rather than the other ($C^-$). This process anchors memory to well-defined genomics concepts. For instance, $C^+$ transitions from being merely a cluster of visual patterns to representing a stable archetype of the omics-defined pattern. This two-fold omics alignment objective, maintaining biological stability through reconstruction while learning separability, ensures the memory is accurate and discriminative.

\paragraph{Knowledge Transfer via WSI Alignment.}
Simultaneously, the aggregated WSI projection, $F_{\text{N-wsi}}$, is forced to align with the omics-calibrated geometry. This forces the WSI encoder to learn modality correlations that are defined by the omics. Since the features from both branch have been normalized: $\|F_{\text{N-wsi}}\|_2 = \|F_{\text{N-omics}}\|_2 =1$, the alignment process can be seen as in a spherical space and the inner product between both modalities with memory, for example, $F_{\text{N-wsi}}^T C^+$ is equivalent to the cosine of the angle between vectors. Based on this, we introduced a soft angle-based 
loss for the alignment $L_{\text{align}}$. Let $\phi(F,C)$ aggregate similarities over the entire memory for any $F \in \{F_{\text{N-wsi}}, F_{\text{N-omics}}\}$:

\begin{equation}
\begin{split}
    \phi(F,C) &= \frac{1}{\tau_{\text{agg}}} \ln \sum_{j=1}^{n} \exp(\tau_{\text{agg}} F^T c_j) \\
    \phi(F,C)&\in\left[-1+\frac{\ln n}{\tau_{\text{agg}}},\ 1+\frac{\ln n}{\tau_{\text{agg}}}\right]
\end{split}
\end{equation}
with the temperature $\tau_{\text{agg}}=5$, the LogSumExp function here smoothly approximates the max similarity while enabling the gradient flow within all memory. We then define the memory differential as:
\begin{equation}
    \Delta(F;C^+,C^-) = \phi(F, C^+) - \phi(F, C^-)\in\left[-2,2\right]
\end{equation}
For a label $y \in \{0,1\}$, the $L_{\text{align}}$ for any $F \in \{F_{\text{N-wsi}}, F_{\text{N-omics}}\}$ is:
\begin{equation}
\begin{aligned}
    &L_{\text{align}}(F, y) = \\
    &\begin{cases}
      \text{softplus}(\beta(\text{margin} - \Delta(F;C^+,C^-))), & \text{if } y=1 \\
      \text{softplus}(\beta(\text{margin} + \Delta(F;C^+,C^-))), & \text{if } y=0
    \end{cases}
\end{aligned}
\end{equation}
with $\beta=20$ as an amplifier factor, and $\text{margin}=0.3$ as a lower-bound of the angle differential between $F$ and $C$.  By enforcing this minimum tolerance, we avoid the ill posed objective of forcing perfect alignment, which could lead to overfitting. The small-batch-stable function pulls $F$ toward the correct memory while pushing it away from false ones, encoding the class boundary in memory geometry without relying on large contrastive batches. Throughout the training process, the memory receives information from both modalities and serves as a semantic bottleneck, forcing the WSI encoder to filter out intra-batch noise and retain only the core semantics relevant to the omics modality.

\paragraph{Memory Evolution via Gradient Decoupling.}
A core challenge is the fundamental asymmetry between modalities: omics data is often a dominant predictor for biomarkers, while WSI features are high-dimensional and sparse. The direct joint-training approach would allow the powerful omics gradient to overwhelm the WSI branch. We therefore decouple branches: there is no direct gradient flow between WSI and omics. Their only interaction is indirect, mediated exclusively by the memory via the alignment loss. Furthermore, we must shield the memory itself from the strong, task-specific gradients of the classification head. Allowing the classifier's loss to backpropagate into the memory would cause memory collapse, as this dominant signal would corrupt the memory being accumulated. In short, the memory evolution is governed by (i) the $L_{\text{mem}}$ which regularizes and maintains orthogonality (introduced in Section~\ref{sec:loss}); (ii) the $L_{\text{align}}$ keeps the memory shaped by meaningful cross-modal alignment; (iii) the reconstruction loss $L_{\text{mse}}$ maintains its internal genomics structure. Even after aggregating all loss terms, their magnitude remains lower than the dominant signal from the classification head. Through the gradient decoupling, we ensure that the memory undergoes a slowly evolving, momentum-like update.

\subsection{Memory-guided Uni-modal Inference}
\label{sec:inference}
During uni-modal inference, the accumulated memory is retrieved.
Each patch-level projection $F_{P\text{-wsi},i}$ is evaluated by its differential affinity to memory representations:
\begin{equation}
    \text{Score}_i = \max_{j} (F_{P\text{-wsi},i}^{\top} c_{j}^{+}) - \max_{j} (F_{P\text{-wsi},i}^{\top} c_{j}^{-}),
\end{equation}
This difference reflects how strongly each patch aligns with the omics-positive visual concept. 
The resulting attention weights are computed with a temperature $\tau=0.2$:
\begin{equation}
\alpha_i = \frac{e^{\text{Score}_i / \tau}}{\sum_{j} e^{\text{Score}_j / \tau}}
\end{equation}
so that patches more consistent with omics-defined patterns receive higher attention. 
The final slide-level representation is obtained as a weighted aggregation: 
\begin{equation}
F_{\text{C-wsi}} = \sum_{i=1}^{I} \alpha_i F_{\text{wsi},i} \in \mathbb{R}^{D}, D=128
\end{equation}
which is subsequently passed through the final linear layer $g(\cdot)$ to generate the slide-level prediction $g(F_{\text{C-wsi}})$. 
According to this, the momentum memory serves as a set of global genomics anchors that guide the attention mechanism, enabling robust predictions.

\subsection{Training Objective}
\label{sec:loss}
The final objective combines cross-entropy loss $L_{\text{ce}}$, reconstruction loss $L_{\text{mse}}$, cross-modal alignment $L_{\text{align}}$, and memory regularization $L_{\text{mem}}$:
\begin{equation}
\begin{split}
    L_{\text{total}} ={}& \lambda_{\text{ce}} L_{\text{ce}}(g(F_{\text{C-wsi}}), y) \\
    & + \lambda_{\text{mse}} L_{\text{mse}}(omics, \text{Decoder}(F_{\text{omics}})) \\
    & + \alpha_{\text{wsi}} L_{\text{align}}(F_{\text{N-wsi}}, y) + \alpha_{\text{omics}} L_{\text{align}}(F_{\text{N-omics}}, y) \\
    & + \lambda_{\text{mem}} L_{\text{mem}}
\end{split}
\end{equation}
Here, $L_{\text{mem}}$ is defined as:
\begin{equation}
\begin{aligned}
    L_{\text{mem}} = & \frac{1}{I} \sum_{i=1}^{I} \| F_{\text{N-wsi},i} - \operatorname{sg}(c_{k^*_i}) \|_2^2 \\
    & +  \sum_{i=1}^{n} \sum_{j \neq i}^{n} \left( \frac{c_i^\top c_j}{\|c_i\|_2 \|c_j\|_2} \right)^2
\end{aligned}
\end{equation}
where $k^*_i = \arg\min_{j} \| F_{\text{N-wsi},i} - c_j \|_2$ denotes the index of the memory closest to the $i$-th patch feature, and $c_{k^*_i}$ represents the corresponding centroid retrieved from the memory bank. The operator $\operatorname{sg}(\cdot)$ indicates the stop-gradient function, which treats its argument as a constant during backpropagation to prevent the memory bank from being updated via direct gradient flow. For the hyperparameters in $L_{\text{total}}$, we apply $\lambda_{\text{ce}}=0.5$, 
$\lambda_{\text{mse}}=0.01$, $\alpha_{\text{omics}}=0.05$, $\alpha_{\text{wsi}}=0.2$, and $\lambda_{\text{mem}}=0.1$. In the supplementary material, we provide the pseudo code for the model. 
\section{Experiments and Results}
\label{sec:experiments}

\begin{table}[t]
    \centering
    \caption{External validation results for the ODX prediction task on the independent in-house dataset. The best performance is in bold, and the second-best performance is underlined.}
    \label{tab:external_results} 
    \begin{tabular}{l ccc}
        \toprule
        \multirow{2}{*}{\textbf{Methods}} & \multicolumn{3}{c}{\textbf{In-house dataset(\%)}} \\
        \cmidrule(lr){2-4}
        & AUC & ACC & F1 \\
        \midrule
        ABMIL   & 75.1$\pm$1.7 & 86.1$\pm$0.2 & 60.9$\pm$3.7 \\
        DSMIL   & 74.3$\pm$2.8 & 86.1$\pm$0.7 & 61.2$\pm$3.2 \\
        TransMIL& 71.7$\pm$2.4 & 85.0$\pm$1.7 & 60.5$\pm$4.1 \\
        DTFDMIL & 76.2$\pm$2.2 & \underline{86.5$\pm$1.5} & \underline{63.5$\pm$3.9} \\
        WIKG    & 75.9$\pm$3.5 & 86.7$\pm$1.4 & 58.3$\pm$5.3 \\
        \midrule
        TDC     & \underline{76.5$\pm$2.1} & 86.2$\pm$3.0 & \underline{63.5$\pm$3.2} \\
        MKD     & 76.2$\pm$2.0 & 86.1$\pm$2.9 & 61.0$\pm$6.3 \\
        G-HANET & 76.1$\pm$1.3 & 86.4$\pm$2.2 & 63.1$\pm$6.4 \\
        \midrule
        \textbf{MoMKD} & \textbf{79.4$\pm$0.8} & \textbf{87.1$\pm$1.7} & \textbf{68.0$\pm$3.0} \\
        \bottomrule
    \end{tabular}
\end{table}

\subsection{Experimental Setup}

\paragraph{Datasets.}
To demonstrate the effectiveness of the proposed QKD method, we conduct extensive experiments on three different tasks based on the TCGA-BRCA dataset \cite{weinsteinCancerGenomeAtlas2013} for HER2, PR, and Oncotype-DX (ODX) score classification. This dataset contains a multi-omics resource. For each task, we build the cohort based on labels provided in the official documentation. The overall label distribution is: HER2: 141 (positive): 668 (negative), PR: 649 (positive): 351 (negative), ODX: 282 (positive): 715 (negative). To evaluate the generalizability of the method, we collected an in-house dataset that provides 1127 breast cancer pathology slides with an ODX score distribution of 162 (positive): 961 (negative). More information for each dataset can be found in the supplementary materials.

\paragraph{Implementation Details.}
For the TCGA-BRCA dataset, we performed patient-level five-fold stratified splitting with the ratio of 7:1:2 for each task and reported the average test metrics across five folds. The in-house dataset serves as the external validation set for the ODX task, which is tested by five models trained on the TCGA-BRCA dataset with the average results reported. The UNIv2~\cite{chenGeneralpurposeFoundationModel2024} is been used as the frozen backbone in this work. More details are available in the supplementary material. For the evaluation, we collected five WSI-only MIL approaches (ABMIL \cite{ilseAttentionbasedDeepMultiple2018}, DSMIL \cite{liDualstreamMultipleInstance2021}, TransMIL \cite{shaoTransMILTransformerBased2021}, DTFDMIL \cite{zhangDTFDMILDoubleTierFeature2022}, WIKG \cite{liDynamicGraphRepresentation2024}) and three multimodal knowledge distillation methods (TDC \cite{xingComprehensiveLearningAdaptive2024}, MKD \cite{zhangMultimodalKnowledgeDecomposition2025}, G-HANet \cite{wangHistoGenomicKnowledgeDistillation2024}) for the comprehensive comparison.

\subsection{Experimental Results}

\paragraph{Internal Comparison.}
Table~\ref{tab:internal_results} shows the results of different classification tasks (HER2, PR, and ODX) on the TCGA-BRCA dataset. As shown in the table, our method consistently outperformed alternative models across all three tasks. Compared with the best-performing WSI-only MIL model (WIKG), our method achieved AUC improvements of $+7.0\%$, $+3.5\%$, and $+5.1\%$ on the HER2, PR, and ODX tasks, respectively. Compared to SOTA multimodal method (MKD), our model still yielded notable gains, achieving AUCs of $79.6\%$, $87.9\%$, and $82.3\%$ for three tasks. These results highlight the effectiveness of the momentum memory distillation design, which enables more discriminative feature alignment between pathology and transcriptomic modalities, thereby improving performance across genomics-sensitive prediction tasks.

\begin{table}[t]
\centering
\caption{Ablation study on the proposed method. All settings are trained on the TCGA-BRCA dataset for the HER2 classification.}
\label{tab:ablation_study}
\resizebox{\columnwidth}{!}{
\begin{tabular}{@{}llc@{}}
\toprule
\textbf{Model Variant} & \textbf{Active Components} & \textbf{AUC (\%)} \\
\midrule
Baseline & WSI solely & 73.9$\pm$3.1 \\
MoMKD ($\alpha_{omics}=0$) & WSI + OMICS Recon + WSI Alignment & 75.2$\pm$2.4 \\
MoMKD ($\alpha_{wsi}=0$) & WSI + OMICS Recon + OMICS Alignment & 75.7$\pm$2.5 \\
MoMKD (w/o Recon) & WSI + Joint Alignment (WSI \& OMICS) & 78.0$\pm$3.6 \\
\textbf{MoMKD (Proposed)} & \textbf{WSI + OMICS Recon + Joint Alignment} & \textbf{79.6$\pm$0.7} \\
\bottomrule
\end{tabular}
}
\end{table}

\paragraph{External Comparison.}
To further assess the generalizability of the proposed MoMKD framework, we conducted experiments on the in-house dataset for the ODX prediction task. As summarized in Table~\ref{tab:external_results}, our method consistently outperformed all competing approaches. Specifically, our method achieved the highest AUC of $79.4\%$, accuracy of $87.1\%$, and F1-score of $68.0\%$, surpassing the best multimodal competitor (TDC, AUC~$= 76.5\%$) by $3.8\%$ in AUC and $7.1\%$ in F1-score. These results confirmed that the accumulated memory alignment enhances cross-domain stability, supporting our hypothesis that momentum-updated memory provides robust data representation for the distribution shift problem.

\begin{figure*}[t]
  \centering
  \includegraphics[width=0.95\textwidth,trim=140 150 150 110, clip]{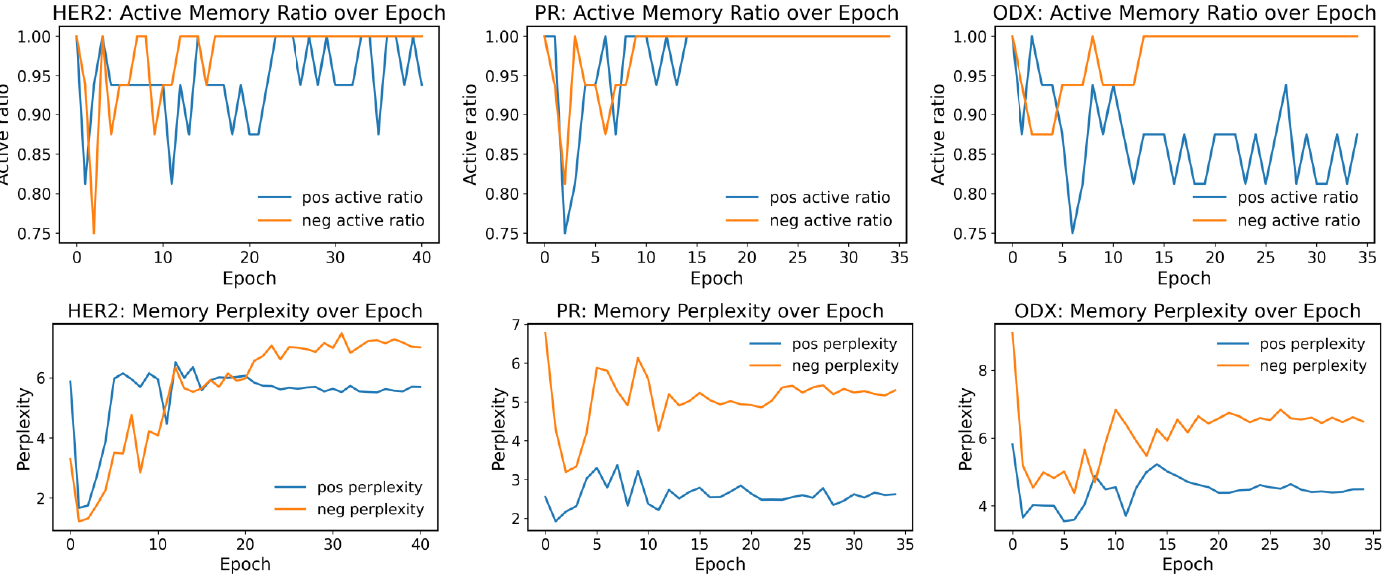}
  \caption{
  Memory dynamics across biomarker prediction tasks.
  Each column corresponds to one task (\textbf{HER2}, \textbf{PR}, and \textbf{ODX}), 
  while each row depicts one statistic of the memory storage: 
  (\textbf{upper}) \emph{Active memory Ratio}, and (\textbf{bottom}) \emph{Perplexity}.
  The active memory ratio 
  quantifies the proportion of memory components utilized during training, 
  ensuring that no dead memory components emerge.
  The corresponding perplexity represents the effective number of active memory components ($c_j^+$ and $c_j^-$).
  Across all tasks, the momentum memory maintains high active ratios ($>0.75$) and perplexity, indicating stable and interpretable utilization of the memory storage.
  }
  \label{fig:codebook_dynamics}
\end{figure*}

\subsection{Ablation Study}
\paragraph{Effects on each component}
To evaluate the distillation performance within the proposed MoMKD framework, we conducted a series of ablation experiments based on the HER2 task, as summarized in Table~\ref{tab:ablation_study}. The WSI-only baseline model achieved an AUC of $73.9 \pm 3.1\%$. First, to isolate the memory's effect as a visual regularizer, we set $\alpha_{\text{omics}}=0$, forcing the memory to be shaped only by WSI features; this configuration alone improved performance to $75.2 \pm 2.4\%$, indicating that the momentum memory can increase the  model's learning capacity. Conversely, incorporating the omics alignment objective alone ($\alpha_{\text{wsi}}=0$), in other words, the omics data actively shapes the geometry of the memory, but the WSI branch only "sees" this genetic-calibrated memory for the final aggregation. This resulted in an AUC of $75.7 \pm 2.5\%$, strongly indicating that the primary performance gain stems from aligning with a stable, genomics-defined memory rather than simple visual consistency. Furthermore, the importance of a stable omics prior was confirmed, as removing the omics self-reconstruction task degraded performance to $78.0 \pm 3.6\%$. Finally, the complete MoMKD framework, jointly optimizing both WSI and omics alignment objectives, achieved the highest performance of $79.6 \pm 0.7\%$, demonstrating the synergistic effect of all components.

\begin{table}[t]
\centering
\caption{Fixed memory vs. momentum memory.}
\label{tab:quantized exp}
\begin{tabular}{@{}lcc@{}}
\toprule
\textbf{Task} & \multicolumn{2}{c}{\textbf{AUC (\%)}} \\ 
\cmidrule(lr){2-3} 
& \textbf{Fixed} & \textbf{Momentum} \\ 
\midrule
HER2 & 75.2$\pm$3.0 & \textbf{79.6$\pm$0.7} \\
PR & 84.7$\pm$2.1 & \textbf{87.9$\pm$0.9} \\
ODX & 81.9$\pm$2.3 & \textbf{82.3$\pm$2.3} \\
\midrule 
In-house dataset & 73.5$\pm$3.7 & \textbf{79.4$\pm$0.8} \\
\bottomrule
\end{tabular}
\end{table}

\paragraph{Fixed memory bank v.s. momentum memory}
To isolate the critical contribution of our dynamic memory update, we compared MoMKD with a fixed memory bank baseline. This baseline uses the exact same K-means initialization, but the memory is subsequently frozen throughout training, reducing it to a static dictionary. As shown in Table~\ref{tab:quantized exp}, our momentum-driven approach yields superior performance across all tasks, including significant gains on HER2 ($+4.5\%$) and the in-house ODX dataset ($+5.9\%$). The crucial limitation of the fixed approach is revealed under domain shift. While the fixed setting achieved a high AUC ($81.1 \pm 2.7\%$) on the public ODX cohort, its performance collapsed to $73.5 \pm 3.7\%$ on the in-house data, indicating severe overfitting to the source domain's initial visual representation. In contrast, MoMKD, with its momentum memory, maintained strong cross-cohort robustness ($79.4 \pm 0.8\%$). This provides clear evidence that the momentum update is essential: it transforms the memory from a static target into a dynamic semantic center. By continuously tracking the global data distribution and smoothing batch-level noise, our method builds a robust semantic representation that generalizes effectively.

\paragraph{Momentum memory update across tasks.}
Fig.~\ref{fig:codebook_dynamics} summarizes the training dynamics of the memory storage across three prediction tasks. 
In all cases, the active memory ratio remains above 0.75, indicating that most memory components participate in the training trajectory. The perplexity profiles show task-specific differences: As the most difficult task, HER2 statues majorly be identified by IHC stain instead of H\&E stain~\cite{HER2TestingBreast,shamaiClinicalUtilityReceptor2024}, it maintains a larger number of active memory components to represent the sparse signal, while PR and ODX converge to lower values since the model can identify more visual signals from the image~\cite{flanaganHistopathologicVariablesPredict2008}.
These observations highlight that momentum memory adapts its effective capacity to task complexity, achieving both stability and compactness.
Detailed definitions of these metrics and additional analyses within the in-house dataset are provided in Supplementary Material Section~S3.

\paragraph{Effects on the memory size}
Fig~S1 in the supplementary material shows the variation in memory size across different tasks, and the MoMKD framework performs consistently.

\begin{figure}[t]
    \centering
    \includegraphics[width=\linewidth,keepaspectratio,
    trim=30 110 350 5, clip]{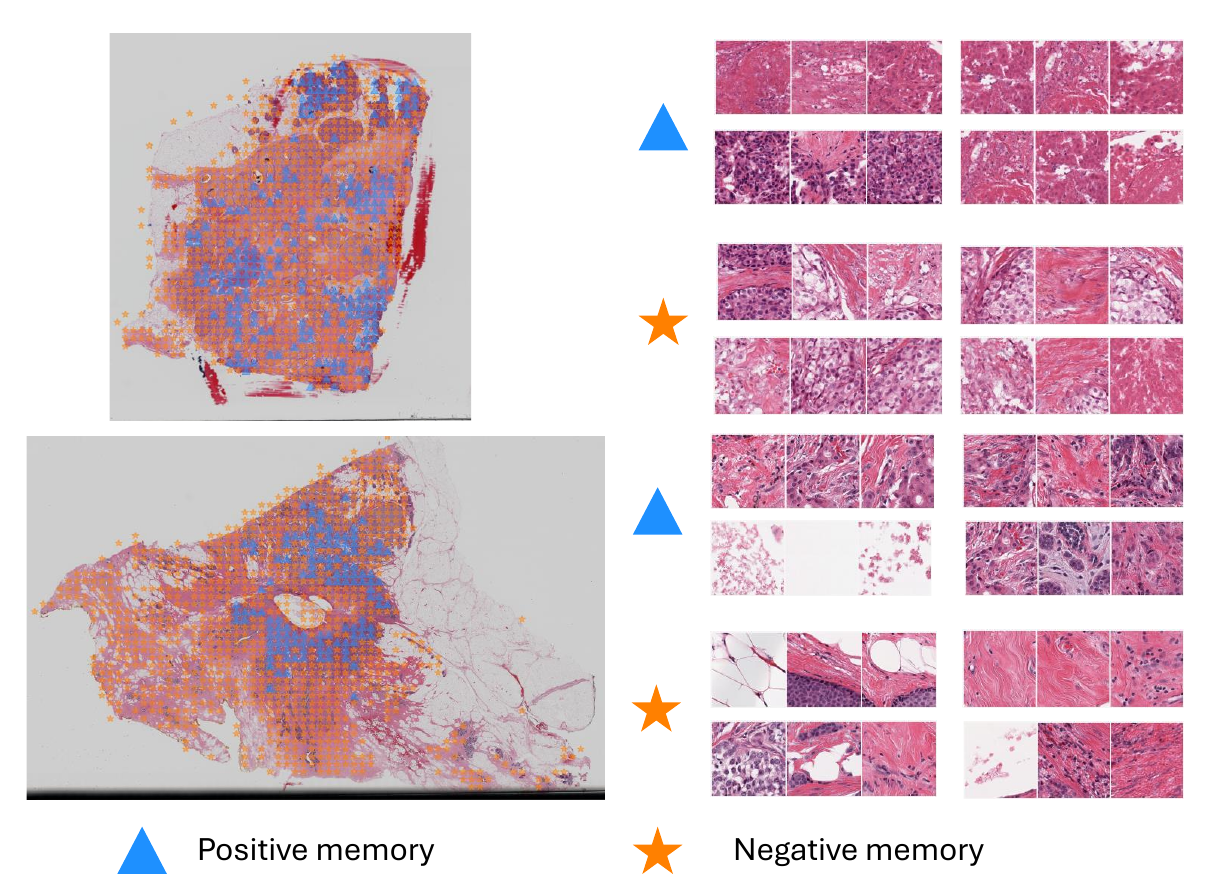} 
    \caption{Visualization result based on memory, with mapped patches in the WSI. The upper figure is the positive case with the lower one being negative based on the Oncotype DX test. For each memory component, we present the top-3 mapped patches here.}
    \label{fig:vis} 
\end{figure}

\subsection{Visualization and Interpretation} 
\label{sec:vis}
To interpret the memory behavior, we visualized the spatial distribution of memory activations on WSIs and examined the corresponding patch-level patterns under the guidance of experienced pathologists. Here we randomly collect four memory components in each memory set with the top three corresponding patches based on the ODX task, as shown in Fig.~\ref{fig:vis}.

Under expert pathological review, it showed the top-ranked image patches (right panel) demonstrate that the positive memory are predominantly activated within tumor-rich and stromal interaction regions, capturing dense epithelial clusters, nuclear pleomorphism, and desmoplastic stroma, which are histopathology consistent with aggressive tumor behavior. In contrast, negative memory responds primarily to benign-appearing structures such as adipose tissue, normal ducts, and fibrous stroma. The top-ranked image patches further confirm that each memory part consistently encodes an interpretable histologic pattern. These findings validate that the learned memory indeed captures biologically meaningful features. Furthermore, an analysis of the misclassified cases (shown in Supplementary Fig. S3) indicates a critical need for more robust filtering methods, as the memory components tend to erroneously concentrate on non-informative white background regions.

\section{Discussion}
\label{sec:discussion}

Overall, our findings demonstrate that the momentum memory plays a central role in stabilizing cross-modal distillation between histopathology and genomics. The memory maintains highly semantic diversity, avoids global collapse, and adapts its usage patterns to the morphological complexity of each biomarker prediction task. Comparisons with static memory further highlight that dynamic, momentum-driven updates are essential for resisting domain shift and constructing a robust semantic representation. These observations suggest that momentum memory is not merely a storage mechanism, but a dynamic semantic aggregator that balances expressivity and compactness. This provides a principled foundation for future multimodal distillation frameworks and opens new possibilities for interpretable and generalizable computational pathology models.

\section{Conclusion}
\label{sec:conclusion}

The cross-modal distillation between histopathology and genomics is hindered by fragile, batch-local alignments. This work presents MoMKD, a Momentum Memory for Knowledge Distillation framework that reframes this problem as alignment to a stable and class-conditioned momentum memory.  Across multiple biomarker prediction tasks (HER2, PR, ODX) on TCGA-BRCA and an independent in-house validation cohort, the proposed method consistently outperforms both WSI-only and multimodal baselines, demonstrating strong generalization and robustness under domain shift. It further indicates that the proposed method successfully bridges the asymmetry between genomics and histopathology modalities. In summary, MoMKD introduces a cross-batch multimodal paradigm for computational pathology, establishing a generalizable principle for stable, interpretable knowledge distillation.

\section{Acknowledgements}
\label{sec:Acknowledgements}
The project described was supported in part by R21 CA273665 (PIs: Gurcan) from the National Cancer Institute, R01 CA276301 (PIs: Niazi, Chen) from the National Cancer Institute, and R21 EB029493 (PIs: Niazi, Segal) from the National Institute of Biomedical Imaging and Bioengineering. The content is solely the responsibility of the authors and does not necessarily represent the official views of the National Institutes of Health, the National Institute of Biomedical Imaging and Bioengineering, or the National Cancer Institute.

{
    \small
    \bibliographystyle{ieeenat_fullname}
    \bibliography{library}

@article{ahmedt-aristizabalSurveyGraphbasedDeep2022,
  title = {A Survey on Graph-Based Deep Learning for Computational Histopathology},
  author = {{Ahmedt-Aristizabal}, David and Armin, Mohammad Ali and Denman, Simon and Fookes, Clinton and Petersson, Lars},
  year = {2022},
  month = jan,
  journal = {Computerized Medical Imaging and Graphics},
  volume = {95},
  pages = {102027},
  issn = {0895-6111},
  doi = {10.1016/j.compmedimag.2021.102027},
  urldate = {2025-10-29}
}

@article{baysoyTechnologicalLandscapeApplications2023,
  title = {The Technological Landscape and Applications of Single-Cell Multi-Omics},
  author = {Baysoy, Alev and Bai, Zhiliang and Satija, Rahul and Fan, Rong},
  year = {2023},
  month = jun,
  journal = {Nature Reviews. Molecular Cell Biology},
  pages = {1--19},
  issn = {1471-0072},
  doi = {10.1038/s41580-023-00615-w},
  urldate = {2025-10-29},
  pmcid = {PMC10242609},
  pmid = {37280296}
}

@article{biMILViTMultipleInstance2023,
  title = {{{MIL-ViT}}: {{A}} Multiple Instance Vision Transformer for Fundus Image Classification},
  shorttitle = {{{MIL-ViT}}},
  author = {Bi, Qi and Sun, Xu and Yu, Shuang and Ma, Kai and Bian, Cheng and Ning, Munan and He, Nanjun and Huang, Yawen and Li, Yuexiang and Liu, Hanruo and Zheng, Yefeng},
  year = {2023},
  month = dec,
  journal = {Journal of Visual Communication and Image Representation},
  volume = {97},
  pages = {103956},
  issn = {1047-3203},
  doi = {10.1016/j.jvcir.2023.103956},
  urldate = {2025-10-29}
}

@article{bontempoGraphBasedMultiScaleApproach2024,
  title = {A {{Graph-Based Multi-Scale Approach With Knowledge Distillation}} for {{WSI Classification}}},
  author = {Bontempo, Gianpaolo and Bolelli, Federico and Porrello, Angelo and Calderara, Simone and Ficarra, Elisa},
  year = {2024},
  month = apr,
  journal = {IEEE Transactions on Medical Imaging},
  volume = {43},
  number = {4},
  pages = {1412--1421},
  issn = {1558-254X},
  doi = {10.1109/TMI.2023.3337549},
  urldate = {2025-11-13}
}

@misc{brodyHowAttentiveAre2022,
  title = {How {{Attentive}} Are {{Graph Attention Networks}}?},
  author = {Brody, Shaked and Alon, Uri and Yahav, Eran},
  year = {2022},
  month = jan,
  number = {arXiv:2105.14491},
  eprint = {2105.14491},
  primaryclass = {cs},
  publisher = {arXiv},
  doi = {10.48550/arXiv.2105.14491},
  urldate = {2025-11-05},
  archiveprefix = {arXiv}
}

@article{chenGeneralpurposeFoundationModel2024,
  title = {Towards a General-Purpose Foundation Model for Computational Pathology},
  author = {Chen, Richard J. and Ding, Tong and Lu, Ming Y. and Williamson, Drew F. K. and Jaume, Guillaume and Song, Andrew H. and Chen, Bowen and Zhang, Andrew and Shao, Daniel and Shaban, Muhammad and Williams, Mane and Oldenburg, Lukas and Weishaupt, Luca L. and Wang, Judy J. and Vaidya, Anurag and Le, Long Phi and Gerber, Georg and Sahai, Sharifa and Williams, Walt and Mahmood, Faisal},
  year = {2024},
  month = mar,
  journal = {Nature Medicine},
  volume = {30},
  number = {3},
  pages = {850--862},
  publisher = {Nature Publishing Group},
  issn = {1546-170X},
  doi = {10.1038/s41591-024-02857-3},
  urldate = {2025-10-17},
  copyright = {2024 The Author(s), under exclusive licence to Springer Nature America, Inc.},
  langid = {english}
}

@article{chenLearningPrivilegedMultimodal2022,
  title = {Learning {{With Privileged Multimodal Knowledge}} for {{Unimodal Segmentation}}},
  author = {Chen, Cheng and Dou, Qi and Jin, Yueming and Liu, Quande and Heng, Pheng Ann},
  year = {2022},
  month = mar,
  journal = {IEEE Transactions on Medical Imaging},
  volume = {41},
  number = {3},
  pages = {621--632},
  issn = {1558-254X},
  doi = {10.1109/TMI.2021.3119385},
  urldate = {2025-10-29}
}

@inproceedings{chenSimpleFrameworkContrastive2020,
  title = {A {{Simple Framework}} for {{Contrastive Learning}} of {{Visual Representations}}},
  booktitle = {Proceedings of the 37th {{International Conference}} on {{Machine Learning}}},
  author = {Chen, Ting and Kornblith, Simon and Norouzi, Mohammad and Hinton, Geoffrey},
  year = {2020},
  month = nov,
  pages = {1597--1607},
  publisher = {PMLR},
  issn = {2640-3498},
  urldate = {2025-02-03},
  langid = {english}
}

@article{flanaganHistopathologicVariablesPredict2008,
  title = {Histopathologic Variables Predict {{Oncotype DX}}™ {{Recurrence Score}}},
  author = {Flanagan, Melina B. and Dabbs, David J. and Brufsky, Adam M. and Beriwal, Sushil and Bhargava, Rohit},
  year = {2008},
  month = oct,
  journal = {Modern Pathology},
  volume = {21},
  number = {10},
  pages = {1255--1261},
  publisher = {Nature Publishing Group},
  issn = {1530-0285},
  doi = {10.1038/modpathol.2008.54},
  urldate = {2025-11-10},
  copyright = {2008 United States and Canadian Academy of Pathology, Inc.},
  langid = {english}
}

@inproceedings{gindraLargeScaleBenchmarkCrossModal2025,
  title = {A {{Large-Scale Benchmark}} of {{Cross-Modal Learning}} for {{Histology}} and {{Gene Expression}} in {{Spatial Transcriptomics}}},
  booktitle = {Proceedings of the {{IEEE}}/{{CVF International Conference}} on {{Computer Vision}}},
  author = {Gindra, Rushin H. and Palla, Giovanni and Nguyen, Mathias and Wagner, Sophia J. and Tran, Manuel and Theis, Fabian J. and Saur, Dieter and Crawford, Lorin and Peng, Tingying},
  year = {2025},
  pages = {1182--1192},
  urldate = {2025-11-13},
  langid = {english}
}

@article{guoBPMambaMILBioinspiredPrototypeguided2025,
  title = {{{BPMambaMIL}}: {{A}} Bio-Inspired Prototype-Guided Multiple Instance Learning for Oncotype {{DX}} Risk Assessment in Histopathology},
  shorttitle = {{{BPMambaMIL}}},
  author = {Guo, Yongxin and Su, Ziyu and Koyun, Onur C. and Lu, Hao and Wesolowski, Robert and Tozbikian, Gary and Niazi, M. Khalid Khan and Gurcan, Metin N.},
  year = {2025},
  month = dec,
  journal = {Computer Methods and Programs in Biomedicine},
  volume = {272},
  pages = {109039},
  issn = {0169-2607},
  doi = {10.1016/j.cmpb.2025.109039},
  urldate = {2025-09-12}
}

@article{gurcanHistopathologicalImageAnalysis2009,
  title = {Histopathological Image Analysis: A Review},
  shorttitle = {Histopathological Image Analysis},
  author = {Gurcan, Metin N. and Boucheron, Laura E. and Can, Ali and Madabhushi, Anant and Rajpoot, Nasir M. and Yener, B.},
  year = {2009},
  journal = {IEEE reviews in biomedical engineering},
  volume = {2},
  pages = {147--171},
  issn = {1941-1189},
  doi = {10.1109/RBME.2009.2034865},
  langid = {english},
  pmcid = {PMC2910932},
  pmid = {20671804}
}

@inproceedings{heMoCoSAMomentumContrast2024,
  title = {{{MoCoSA}}: {{Momentum Contrast}} for {{Knowledge Graph Completion}} with {{Structure-Augmented Pre-trained Language Models}}},
  shorttitle = {{{MoCoSA}}},
  booktitle = {2024 {{IEEE International Conference}} on {{Multimedia}} and {{Expo}} ({{ICME}})},
  author = {He, Jiabang and Liu, Jia and Wang, Lei and Li, Xiyao and Xu, Xing},
  year = {2024},
  month = jul,
  pages = {1--6},
  issn = {1945-788X},
  doi = {10.1109/ICME57554.2024.10687798},
  urldate = {2025-11-13}
}

@misc{heMomentumContrastUnsupervised2020,
  title = {Momentum {{Contrast}} for {{Unsupervised Visual Representation Learning}}},
  author = {He, Kaiming and Fan, Haoqi and Wu, Yuxin and Xie, Saining and Girshick, Ross},
  year = {2020},
  month = mar,
  number = {arXiv:1911.05722},
  eprint = {1911.05722},
  primaryclass = {cs},
  publisher = {arXiv},
  doi = {10.48550/arXiv.1911.05722},
  urldate = {2025-11-13},
  archiveprefix = {arXiv}
}

@misc{HER2TestingBreast,
  title = {{{HER2 Testing}} in {{Breast Cancer}} - 2023 {{Guideline Update}}},
  journal = {College of American Pathologists},
  urldate = {2025-11-10},
  howpublished = {https://www.cap.org/protocols-and-guidelines/cap-guidelines/current-cap-guidelines/recommendations-for-human-epidermal-growth-factor-2-testing-in-breast-cancer},
  langid = {american}
}

@misc{hintonDistillingKnowledgeNeural2015,
  title = {Distilling the {{Knowledge}} in a {{Neural Network}}},
  author = {Hinton, Geoffrey and Vinyals, Oriol and Dean, Jeff},
  year = {2015},
  month = mar,
  number = {arXiv:1503.02531},
  eprint = {1503.02531},
  primaryclass = {stat},
  publisher = {arXiv},
  doi = {10.48550/arXiv.1503.02531},
  urldate = {2025-10-29},
  archiveprefix = {arXiv}
}

@article{howardIntegrationClinicalFeatures2023,
  title = {Integration of Clinical Features and Deep Learning on Pathology for the Prediction of Breast Cancer Recurrence Assays and Risk of Recurrence},
  author = {Howard, Frederick M. and Dolezal, James and Kochanny, Sara and Khramtsova, Galina and Vickery, Jasmine and Srisuwananukorn, Andrew and Woodard, Anna and Chen, Nan and Nanda, Rita and Perou, Charles M. and Olopade, Olufunmilayo I. and Huo, Dezheng and Pearson, Alexander T.},
  year = {2023},
  month = apr,
  journal = {npj Breast Cancer},
  volume = {9},
  number = {1},
  pages = {1--6},
  publisher = {Nature Publishing Group},
  issn = {2374-4677},
  doi = {10.1038/s41523-023-00530-5},
  urldate = {2024-12-16},
  copyright = {2023 The Author(s)},
  langid = {english}
}

@inproceedings{ilseAttentionbasedDeepMultiple2018,
  title = {Attention-Based {{Deep Multiple Instance Learning}}},
  booktitle = {Proceedings of the 35th {{International Conference}} on {{Machine Learning}}},
  author = {Ilse, Maximilian and Tomczak, Jakub and Welling, Max},
  year = {2018},
  month = jul,
  pages = {2127--2136},
  publisher = {PMLR},
  issn = {2640-3498},
  urldate = {2025-01-14},
  langid = {english}
}

@article{ivanovaStandardizedPathologyReport2024,
  title = {Standardized Pathology Report for {{HER2}} Testing in Compliance with 2023 {{ASCO}}/{{CAP}} Updates and 2023 {{ESMO}} Consensus Statements on {{HER2-low}} Breast Cancer},
  author = {Ivanova, Mariia and Porta, Francesca Maria and D'Ercole, Marianna and Pescia, Carlo and Sajjadi, Elham and Cursano, Giulia and De Camilli, Elisa and Pala, Oriana and Mazzarol, Giovanni and Venetis, Konstantinos and {Guerini-Rocco}, Elena and Curigliano, Giuseppe and Viale, Giuseppe and Fusco, Nicola},
  year = {2024},
  journal = {Virchows Archiv},
  volume = {484},
  number = {1},
  pages = {3--14},
  issn = {0945-6317},
  doi = {10.1007/s00428-023-03656-w},
  urldate = {2025-11-10},
  pmcid = {PMC10791807},
  pmid = {37770765}
}

@inproceedings{liDualstreamMultipleInstance2021,
  title = {Dual-Stream Multiple Instance Learning Network for Whole Slide Image Classification with Self-Supervised Contrastive Learning},
  booktitle = {Proceedings of the {{IEEE}}/{{CVF}} Conference on Computer Vision and Pattern Recognition},
  author = {Li, Bin and Li, Yin and Eliceiri, Kevin W.},
  year = {2021},
  pages = {14318--14328},
  urldate = {2025-01-31}
}

@inproceedings{liDynamicGraphRepresentation2024,
  title = {Dynamic {{Graph Representation}} with {{Knowledge-Aware Attention}} for {{Histopathology Whole Slide Image Analysis}}},
  booktitle = {2024 {{IEEE}}/{{CVF Conference}} on {{Computer Vision}} and {{Pattern Recognition}} ({{CVPR}})},
  author = {Li, Jiawen and Chen, Yuxuan and Chu, Hongbo and Sun, Qiehe and Guan, Tian and Han, Anjia and He, Yonghong},
  year = {2024},
  month = jun,
  pages = {11323--11332},
  publisher = {IEEE},
  address = {Seattle, WA, USA},
  doi = {10.1109/CVPR52733.2024.01076},
  urldate = {2025-10-29},
  copyright = {https://doi.org/10.15223/policy-029},
  isbn = {9798350353006},
  langid = {english}
}

@article{luDataefficientWeaklySupervised2021a,
  title = {Data-Efficient and Weakly Supervised Computational Pathology on Whole-Slide Images},
  author = {Lu, Ming Y. and Williamson, Drew F. K. and Chen, Tiffany Y. and Chen, Richard J. and Barbieri, Matteo and Mahmood, Faisal},
  year = {2021},
  month = jun,
  journal = {Nature Biomedical Engineering},
  volume = {5},
  number = {6},
  pages = {555--570},
  publisher = {Nature Publishing Group},
  issn = {2157-846X},
  doi = {10.1038/s41551-020-00682-w},
  urldate = {2025-10-29},
  copyright = {2021 The Author(s), under exclusive licence to Springer Nature Limited},
  langid = {english}
}

@article{maGeneralizablePathologyFoundation2025,
  title = {A Generalizable Pathology Foundation Model Using a Unified Knowledge Distillation Pretraining Framework},
  author = {Ma, Jiabo and Guo, Zhengrui and Zhou, Fengtao and Wang, Yihui and Xu, Yingxue and Li, Jinbang and Yan, Fang and Cai, Yu and Zhu, Zhengjie and Jin, Cheng and Lin, Yi and Jiang, Xinrui and Zhao, Chenglong and Li, Danyi and Han, Anjia and Li, Zhenhui and Chan, Ronald Cheong Kin and Wang, Jiguang and Fei, Peng and Cheng, Kwang-Ting and Zhang, Shaoting and Liang, Li and Chen, Hao},
  year = {2025},
  month = sep,
  journal = {Nature Biomedical Engineering},
  pages = {1--20},
  publisher = {Nature Publishing Group},
  issn = {2157-846X},
  doi = {10.1038/s41551-025-01488-4},
  urldate = {2025-10-29},
  copyright = {2025 The Author(s), under exclusive licence to Springer Nature Limited},
  langid = {english}
}

@article{nemaOmicsbasedTumorMicroenvironment2024,
  title = {An Omics-Based Tumor Microenvironment Approach and Its Prospects},
  author = {Nema, Rajeev},
  year = {2024},
  journal = {Reports of Practical Oncology and Radiotherapy},
  volume = {29},
  number = {5},
  pages = {649--650},
  issn = {2083-4640},
  doi = {10.5603/rpor.102823},
  urldate = {2025-10-29},
  copyright = {Completion of the online submission form electronically (the Author's Statement) is tantamount to automatic transfer of the copyright for publishing and distribution of the submitted material (in all current and future forms and fields of exploitation) to the Copyright Owner, i.e. Greater Poland Cancer Centre (Poznan, Poland), on condition that these materials are accepted for publication. The authors agree not to publish any data or figures presented in their work in any place or in any language without the prior written consent of the Publisher.},
  langid = {english}
}

@article{paikMultigeneAssayPredict2004,
  title = {A {{Multigene Assay}} to {{Predict Recurrence}} of {{Tamoxifen-Treated}}, {{Node-Negative Breast Cancer}}},
  author = {Paik, Soonmyung and Shak, Steven and Tang, Gong and Kim, Chungyeul and Baker, Joffre and Cronin, Maureen and Baehner, Frederick L. and Walker, Michael G. and Watson, Drew and Park, Taesung and Hiller, William and Fisher, Edwin R. and Wickerham, D. Lawrence and Bryant, John and Wolmark, Norman},
  year = {2004},
  month = dec,
  journal = {New England Journal of Medicine},
  volume = {351},
  number = {27},
  pages = {2817--2826},
  issn = {0028-4793, 1533-4406},
  doi = {10.1056/NEJMoa041588},
  urldate = {2025-01-13},
  langid = {english}
}

@article{shamaiClinicalUtilityReceptor2024,
  title = {Clinical Utility of Receptor Status Prediction in Breast Cancer and Misdiagnosis Identification Using Deep Learning on Hematoxylin and Eosin-Stained Slides},
  author = {Shamai, Gil and Schley, Ran and Cretu, Alexandra and Neoran, Tal and Sabo, Edmond and Binenbaum, Yoav and Cohen, Shachar and Goldman, Tal and Pol{\'o}nia, Ant{\'o}nio and Drumea, Keren and Stoliar, Karin and Kimmel, Ron},
  year = {2024},
  month = dec,
  journal = {Communications Medicine},
  volume = {4},
  number = {1},
  pages = {276},
  publisher = {Nature Publishing Group},
  issn = {2730-664X},
  doi = {10.1038/s43856-024-00695-5},
  urldate = {2025-11-10},
  copyright = {2025 The Author(s)},
  langid = {english}
}

@inproceedings{shaoTransMILTransformerBased2021,
  title = {{{TransMIL}}: {{Transformer}} Based {{Correlated Multiple Instance Learning}} for {{Whole Slide Image Classification}}},
  shorttitle = {{{TransMIL}}},
  booktitle = {Advances in {{Neural Information Processing Systems}}},
  author = {Shao, Zhuchen and Bian, Hao and Chen, Yang and Wang, Yifeng and Zhang, Jian and Ji, Xiangyang and {zhang}, yongbing},
  year = {2021},
  volume = {34},
  pages = {2136--2147},
  publisher = {Curran Associates, Inc.},
  urldate = {2025-01-14}
}

@inproceedings{shuSlideGCDSlideBasedGraph2024,
  title = {{{SlideGCD}}: {{Slide-Based Graph Collaborative Training}} with~{{Knowledge Distillation}} for~{{Whole Slide Image Classification}}},
  shorttitle = {{{SlideGCD}}},
  booktitle = {Medical {{Image Computing}} and {{Computer Assisted Intervention}} -- {{MICCAI}} 2024},
  author = {Shu, Tong and Shi, Jun and Sun, Dongdong and Jiang, Zhiguo and Zheng, Yushan},
  editor = {Linguraru, Marius George and Dou, Qi and Feragen, Aasa and Giannarou, Stamatia and Glocker, Ben and Lekadir, Karim and Schnabel, Julia A.},
  year = {2024},
  pages = {470--480},
  publisher = {Springer Nature Switzerland},
  address = {Cham},
  doi = {10.1007/978-3-031-72083-3_44},
  isbn = {978-3-031-72083-3},
  langid = {english}
}

@article{suBCRNetDeepLearning2023,
  title = {{{BCR-Net}}: {{A}} Deep Learning Framework to Predict Breast Cancer Recurrence from Histopathology Images},
  shorttitle = {{{BCR-Net}}},
  author = {Su, Ziyu and Niazi, Muhammad Khalid Khan and Tavolara, Thomas E. and Niu, Shuo and Tozbikian, Gary H. and Wesolowski, Robert and Gurcan, Metin N.},
  year = {2023},
  journal = {PloS One},
  volume = {18},
  number = {4},
  pages = {e0283562},
  issn = {1932-6203},
  doi = {10.1371/journal.pone.0283562},
  langid = {english},
  pmcid = {PMC10072418},
  pmid = {37014891}
}

@misc{suComputationalPathologyAccurate2024,
  title = {Computational {{Pathology}} for {{Accurate Prediction}} of {{Breast Cancer Recurrence}}: {{Development}} and {{Validation}} of a {{Deep Learning-based Tool}}},
  shorttitle = {Computational {{Pathology}} for {{Accurate Prediction}} of {{Breast Cancer Recurrence}}},
  author = {Su, Ziyu and Guo, Yongxin and Wesolowski, Robert and Tozbikian, Gary and O'Connell, Nathaniel S. and Niazi, M. Khalid Khan and Gurcan, Metin N.},
  year = {2024},
  month = sep,
  number = {arXiv:2409.15491},
  eprint = {2409.15491},
  primaryclass = {eess},
  publisher = {arXiv},
  doi = {10.48550/arXiv.2409.15491},
  urldate = {2025-01-05},
  archiveprefix = {arXiv}
}

@misc{suComputationalPathologyAccurate2024b,
  title = {Computational {{Pathology}} for {{Accurate Prediction}} of {{Breast Cancer Recurrence}}: {{Development}} and {{Validation}} of a {{Deep Learning-based Tool}}},
  shorttitle = {Computational {{Pathology}} for {{Accurate Prediction}} of {{Breast Cancer Recurrence}}},
  author = {Su, Ziyu and Guo, Yongxin and Wesolowski, Robert and Tozbikian, Gary and O'Connell, Nathaniel S. and Niazi, M. Khalid Khan and Gurcan, Metin N.},
  year = {2024},
  month = sep,
  number = {arXiv:2409.15491},
  eprint = {2409.15491},
  primaryclass = {eess},
  publisher = {arXiv},
  doi = {10.48550/arXiv.2409.15491},
  urldate = {2025-11-10},
  archiveprefix = {arXiv}
}

@misc{suComputationalPathologyAccurate2024c,
  title = {Computational {{Pathology}} for {{Accurate Prediction}} of {{Breast Cancer Recurrence}}: {{Development}} and {{Validation}} of a {{Deep Learning-based Tool}}},
  shorttitle = {Computational {{Pathology}} for {{Accurate Prediction}} of {{Breast Cancer Recurrence}}},
  author = {Su, Ziyu and Guo, Yongxin and Wesolowski, Robert and Tozbikian, Gary and O'Connell, Nathaniel S. and Niazi, M. Khalid Khan and Gurcan, Metin N.},
  year = {2024},
  month = sep,
  number = {arXiv:2409.15491},
  eprint = {2409.15491},
  primaryclass = {eess},
  publisher = {arXiv},
  doi = {10.48550/arXiv.2409.15491},
  urldate = {2025-11-10},
  archiveprefix = {arXiv}
}

@article{valierisWeaklysupervisedDeepLearning2024,
  title = {Weakly-Supervised Deep Learning Models Enable {{HER2-low}} Prediction from {{H}} \&{{E}} Stained Slides},
  author = {Valieris, Renan and Martins, Luan and Defelicibus, Alexandre and Bueno, Adriana Passos and {de Toledo Osorio}, Cynthia Aparecida Bueno and Carraro, Dirce and {Dias-Neto}, Emmanuel and Rosales, Rafael A. and {de Figueiredo}, Jose Marcio Barros and da Silva, Israel Tojal},
  year = {2024},
  month = aug,
  journal = {Breast Cancer Research},
  volume = {26},
  number = {1},
  pages = {124},
  issn = {1465-542X},
  doi = {10.1186/s13058-024-01863-0},
  urldate = {2025-11-10}
}

@article{wangDistillingHeterogeneousKnowledge2025,
  title = {Distilling Heterogeneous Knowledge with Aligned Biological Entities for Histological Image Classification},
  author = {Wang, Kang and Zheng, Feiyang and Guan, Dayan and Liu, Jia and Qin, Jing},
  year = {2025},
  month = apr,
  journal = {Pattern Recognition},
  volume = {160},
  pages = {111173},
  issn = {0031-3203},
  doi = {10.1016/j.patcog.2024.111173},
  urldate = {2025-11-13}
}

@misc{wangHistoGenomicKnowledgeDistillation2024,
  title = {Histo-{{Genomic Knowledge Distillation For Cancer Prognosis From Histopathology Whole Slide Images}}},
  author = {Wang, Zhikang and Zhang, Yumeng and Xu, Yingxue and Imoto, Seiya and Chen, Hao and Song, Jiangning},
  year = {2024},
  month = mar,
  number = {arXiv:2403.10040},
  eprint = {2403.10040},
  primaryclass = {eess},
  publisher = {arXiv},
  doi = {10.48550/arXiv.2403.10040},
  urldate = {2025-04-20},
  archiveprefix = {arXiv}
}

@article{weinsteinCancerGenomeAtlas2013,
  title = {The {{Cancer Genome Atlas Pan-Cancer}} Analysis Project},
  author = {Weinstein, John N. and Collisson, Eric A. and Mills, Gordon B. and Shaw, Kenna R. Mills and Ozenberger, Brad A. and Ellrott, Kyle and Shmulevich, Ilya and Sander, Chris and Stuart, Joshua M.},
  year = {2013},
  month = oct,
  journal = {Nature Genetics},
  volume = {45},
  number = {10},
  pages = {1113--1120},
  publisher = {Nature Publishing Group},
  issn = {1546-1718},
  doi = {10.1038/ng.2764},
  urldate = {2025-11-10},
  copyright = {2013 The Author(s)},
  langid = {english}
}

@article{whitneyQuantitativeNuclearHistomorphometry2018a,
  title = {Quantitative Nuclear Histomorphometry Predicts Oncotype {{DX}} Risk Categories for Early Stage {{ER}}+ Breast Cancer},
  author = {Whitney, Jon and Corredor, German and Janowczyk, Andrew and Ganesan, Shridar and Doyle, Scott and Tomaszewski, John and Feldman, Michael and Gilmore, Hannah and Madabhushi, Anant},
  year = {2018},
  month = may,
  journal = {BMC Cancer},
  volume = {18},
  number = {1},
  pages = {610},
  issn = {1471-2407},
  doi = {10.1186/s12885-018-4448-9},
  urldate = {2025-11-10}
}

@inproceedings{wuUnsupervisedFeatureLearning2018,
  title = {Unsupervised {{Feature Learning}} via {{Non-Parametric Instance Discrimination}}},
  booktitle = {Proceedings of the {{IEEE Conference}} on {{Computer Vision}} and {{Pattern Recognition}}},
  author = {Wu, Zhirong and Xiong, Yuanjun and Yu, Stella X. and Lin, Dahua},
  year = {2018},
  pages = {3733--3742},
  urldate = {2025-11-13}
}

@article{xingComprehensiveLearningAdaptive2024,
  title = {Comprehensive Learning and Adaptive Teaching: {{Distilling}} Multi-Modal Knowledge for Pathological Glioma Grading},
  shorttitle = {Comprehensive Learning and Adaptive Teaching},
  author = {Xing, Xiaohan and Zhu, Meilu and Chen, Zhen and Yuan, Yixuan},
  year = {2024},
  month = jan,
  journal = {Medical Image Analysis},
  volume = {91},
  pages = {102990},
  issn = {1361-8415},
  doi = {10.1016/j.media.2023.102990},
  urldate = {2025-09-15}
}

@article{yuMultimodalDataFusion2024,
  title = {Multimodal Data Fusion {{AI}} Model Uncovers Tumor Microenvironment Immunotyping Heterogeneity and Enhanced Risk Stratification of Breast Cancer},
  author = {Yu, Yunfang and Cai, Gengyi and Lin, Ruichong and Wang, Zehua and Chen, Yongjian and Tan, Yujie and He, Zifan and Sun, Zhuo and Ouyang, Wenhao and Yao, Herui and Zhang, Kang},
  year = {2024},
  journal = {MedComm},
  volume = {5},
  number = {12},
  pages = {e70023},
  issn = {2688-2663},
  doi = {10.1002/mco2.70023},
  urldate = {2025-10-29},
  copyright = {{\copyright} 2024 The Author(s). MedComm published by Sichuan International Medical Exchange \& Promotion Association (SCIMEA) and John Wiley \& Sons Australia, Ltd.},
  langid = {english}
}

@misc{zhangAcceleratingDataProcessing2025,
  title = {Accelerating {{Data Processing}} and {{Benchmarking}} of {{AI Models}} for {{Pathology}}},
  author = {Zhang, Andrew and Jaume, Guillaume and Vaidya, Anurag and Ding, Tong and Mahmood, Faisal},
  year = {2025},
  month = feb,
  number = {arXiv:2502.06750},
  eprint = {2502.06750},
  primaryclass = {cs},
  publisher = {arXiv},
  doi = {10.48550/arXiv.2502.06750},
  urldate = {2025-10-17},
  archiveprefix = {arXiv}
}

@misc{zhangDisentangledMultimodalLearning2025,
  title = {Disentangled {{Multi-modal Learning}} of {{Histology}} and {{Transcriptomics}} for {{Cancer Characterization}}},
  author = {Zhang, Yupei and Wang, Xiaofei and Liu, Anran and Yu, Lequan and Li, Chao},
  year = {2025},
  month = aug,
  number = {arXiv:2508.16479},
  eprint = {2508.16479},
  primaryclass = {eess},
  publisher = {arXiv},
  doi = {10.48550/arXiv.2508.16479},
  urldate = {2025-11-05},
  archiveprefix = {arXiv}
}

@misc{zhangDTFDMILDoubleTierFeature2022,
  title = {{{DTFD-MIL}}: {{Double-Tier Feature Distillation Multiple Instance Learning}} for {{Histopathology Whole Slide Image Classification}}},
  shorttitle = {{{DTFD-MIL}}},
  author = {Zhang, Hongrun and Meng, Yanda and Zhao, Yitian and Qiao, Yihong and Yang, Xiaoyun and Coupland, Sarah E. and Zheng, Yalin},
  year = {2022},
  month = mar,
  number = {arXiv:2203.12081},
  eprint = {2203.12081},
  primaryclass = {cs},
  publisher = {arXiv},
  doi = {10.48550/arXiv.2203.12081},
  urldate = {2025-02-10},
  archiveprefix = {arXiv}
}

@misc{zhangMultimodalKnowledgeDecomposition2025,
  title = {Multi-Modal {{Knowledge Decomposition}} Based {{Online Distillation}} for {{Biomarker Prediction}} in {{Breast Cancer Histopathology}}},
  author = {Zhang, Qibin and Hao, Xinyu and Chen, Qiao and Xu, Rui and Cong, Fengyu and Lu, Cheng and Xu, Hongming},
  year = {2025},
  month = aug,
  number = {arXiv:2508.17213},
  eprint = {2508.17213},
  primaryclass = {cs},
  publisher = {arXiv},
  doi = {10.48550/arXiv.2508.17213},
  urldate = {2025-09-09},
  archiveprefix = {arXiv}
}

@article{zhangUSERUnifiedSemantic2024,
  title = {{{USER}}: {{Unified Semantic Enhancement With Momentum Contrast}} for {{Image-Text Retrieval}}},
  shorttitle = {{{USER}}},
  author = {Zhang, Yan and Ji, Zhong and Wang, Di and Pang, Yanwei and Li, Xuelong},
  year = {2024},
  journal = {IEEE Transactions on Image Processing},
  volume = {33},
  pages = {595--609},
  issn = {1941-0042},
  doi = {10.1109/TIP.2023.3348297},
  urldate = {2025-11-13}
}

@inproceedings{zhuDGRMILExploringDiverse2025,
  title = {{{DGR-MIL}}: {{Exploring Diverse Global Representation}} in~{{Multiple Instance Learning}} for~{{Whole Slide Image Classification}}},
  shorttitle = {{{DGR-MIL}}},
  booktitle = {Computer {{Vision}} -- {{ECCV}} 2024},
  author = {Zhu, Wenhui and Chen, Xiwen and Qiu, Peijie and Sotiras, Aristeidis and Razi, Abolfazl and Wang, Yalin},
  editor = {Leonardis, Ale{\v s} and Ricci, Elisa and Roth, Stefan and Russakovsky, Olga and Sattler, Torsten and Varol, G{\"u}l},
  year = {2025},
  pages = {333--351},
  publisher = {Springer Nature Switzerland},
  address = {Cham},
  doi = {10.1007/978-3-031-72920-1_19},
  isbn = {978-3-031-72920-1},
  langid = {english}
}

@incollection{chen2021whole,
  title={Whole Slide Images are 2D Point Clouds: Context-Aware Survival Prediction using Patch-based Graph Convolutional Networks},
  author={Chen, Richard J and Lu, Ming Y and Shaban, Muhammad and Chen, Chengkuan and Chen, Tiffany Y and Williamson, Drew FK and Mahmood, Faisal},
  doi = {10.1007/978-3-030-87237-3_33},
  url = {https://doi.org/10.1007/978-3-030-87237-3_33},
  publisher = {Springer International Publishing},
  pages = {339--349},
  booktitle = {Medical Image Computing and Computer Assisted Intervention {\textendash} {MICCAI} 2021},
  year = {2021}
}

@article{2f3f68f41271474c8e5638b13b6b8f8f,
title = "Unsupervised Hyperspectral Image Super-Resolution via Self-Supervised Modality Decoupling",
author = "Songcheng Du and Yang Zou and Zixu Wang and Xingyuan Li and Ying Li and Changjing Shang and Qiang Shen",
year = "2026",
month = jan,
day = "9",
language = "English",
journal = "International Journal of Computer Vision",
issn = "0920-5691",
publisher = "Springer Nature",
}

@misc{lan2025acamkd,
      title={ACAM-KD: Adaptive and Cooperative Attention Masking for Knowledge Distillation}, 
      author={Qizhen Lan and Qing Tian},
      year={2025},
      eprint={2503.06307},
      archivePrefix={arXiv},
      primaryClass={cs.CV},
      url={https://arxiv.org/abs/2503.06307}, 
}

@misc{lan2026recokd,
      title={ReCo-KD: Region- and Context-Aware Knowledge Distillation for Efficient 3D Medical Image Segmentation}, 
      author={Qizhen Lan and Yu-Chun Hsu and Nida Saddaf Khan and Xiaoqian Jiang},
      year={2026},
      eprint={2603.06307}, 
      archivePrefix={arXiv},
      primaryClass={eess.IV},
      url={https://arxiv.org/abs/2603.06307},
}

@article{sun2025hyperpoint,
  title={HyperPoint: Multimodal 3D foundation model in hyperbolic space},
  author={Sun, Yiding and Cheng, Haozhe and Lu, Chaoyi and Li, Zhengqiao and Wu, Minghong and Lu, Huimin and Zhu, Jihua},
  journal={Pattern Recognition},
  pages={112800},
  year={2025},
  publisher={Elsevier}
}

@misc{su2025medgrpomultitaskreinforcementlearning,
      title={MedGRPO: Multi-Task Reinforcement Learning for Heterogeneous Medical Video Understanding}, 
      author={Yuhao Su and Anwesa Choudhuri and Zhongpai Gao and Benjamin Planche and Van Nguyen Nguyen and Meng Zheng and Yuhan Shen and Arun Innanje and Terrence Chen and Ehsan Elhamifar and Ziyan Wu},
      year={2025},
      eprint={2512.06581},
      archivePrefix={arXiv},
      primaryClass={cs.CV},
      url={https://arxiv.org/abs/2512.06581}, 
}

@inproceedings{li2025pointdico,
  title={PointDico: Contrastive 3D Representation Learning Guided by Diffusion Models},
  author={Li, Pengbo and Sun, Yiding and Cheng, Haozhe},
  booktitle={2025 International Joint Conference on Neural Networks (IJCNN)},
  pages={1--9},
  year={2025},
  organization={IEEE}
}

@ARTICLE{10530449,
  author={Li, Qiankun and Wang, Yimou and Zhang, Yani and Zuo, Zhaoyu and Chen, Junxin and Wang, Wei},
  journal={IEEE Transactions on Fuzzy Systems}, 
  title={Fuzzy-ViT: A Deep Neuro-Fuzzy System for Cross-Domain Transfer Learning From Large-Scale General Data to Medical Image}, 
  year={2025},
  volume={33},
  number={1},
  pages={231-241},
  doi={10.1109/TFUZZ.2024.3400861}}

@article{li2025unleashing,
  title={Unleashing Foundation Vision Models: Adaptive Transfer for Diverse Data-Limited Scientific Domains},
  author={Li, Qiankun and He, Feng and Chen, Huabao and Ning, Xin and Wang, Kun and Wang, Zengfu},
  journal={arXiv preprint arXiv:2512.22664},
  year={2025}
}

@misc{gong2025medcmr,
      title={Med-CMR: A Fine-Grained Benchmark Integrating Visual Evidence and Clinical Logic for Medical Complex Multimodal Reasoning}, 
      author={Haozhen Gong and Xiaozhong Ji and Yuansen Liu and Wenbin Wu and Xiaoxiao Yan and Jingjing Liu and Kai Wu and Jiazhen Pan and Bailiang Jian and Jiangning Zhang and Xiaobin Hu and Hongwei Bran Li},
      year={2025},
      eprint={2512.00818},
      archivePrefix={arXiv},
      primaryClass={cs.AI},
      url={https://arxiv.org/abs/2512.00818}, 
}

@inproceedings{chen2021multimodal,
  title={Multimodal co-attention transformer for survival prediction in gigapixel whole slide images},
  author={Chen, Richard J and Lu, Ming Y and Weng, Wei-Hung and Chen, Tiffany Y and Williamson, Drew FK and Manz, Trevor and Shady, Maha and Mahmood, Faisal},
  booktitle={Proceedings of the IEEE/CVF international conference on computer vision},
  pages={4015--4025},
  year={2021}
}

@article{chen2022pan,
  title={Pan-cancer integrative histology-genomic analysis via multimodal deep learning},
  author={Chen, Richard J and Lu, Ming Y and Williamson, Drew FK and Chen, Tiffany Y and Lipkova, Jana and Noor, Zahra and Shaban, Muhammad and Shady, Maha and Williams, Mane and Joo, Bumjin and others},
  journal={Cancer cell},
  volume={40},
  number={8},
  pages={865--878},
  year={2022},
  publisher={Elsevier}
}

@InProceedings{Xu_2023_ICCV,
    author    = {Xu, Yingxue and Chen, Hao},
    title     = {Multimodal Optimal Transport-based Co-Attention Transformer with Global Structure Consistency for Survival Prediction},
    booktitle = {Proceedings of the IEEE/CVF International Conference on Computer Vision (ICCV)},
    month     = {October},
    year      = {2023},
    pages     = {21241-21251}
}

@String(CVPR= {IEEE Conf. Comput. Vis. Pattern Recog.})

@String(ICCV= {Int. Conf. Comput. Vis.})

@String(ECCV= {Eur. Conf. Comput. Vis.})

@String(ICME = {Int. Conf. Multimedia and Expo})

@String(CVPR  = {CVPR})

@String(ICCV  = {ICCV})

@String(ECCV  = {ECCV})

@String(ICME  =	{ICME})
}

\end{document}


\title{Momentum Memory for Knowledge Distillation in Computational Pathology}

\author{
    Yongxin Guo$^{1}$\thanks{Corresponding author: Yongxin.Guo@wfusm.edu.\\ Code: \url{https://github.com/CAIR-LAB-WFUSM/MoMKD}}, \ 
    Hao Lu$^{1}$, \ 
    Onur C. Koyun$^{1}$, \ 
    Zhengjie Zhu$^{1}$, \ 
    Muhammet F. Demir$^{1}$, \ 
    Metin N. Gurcan$^{1}$ \\[4pt]
    $^{1}$Wake Forest University School of Medicine, Winston-Salem, NC, USA \\[4pt]
    {\tt\small \{Hao.Lu, Onur.Koyun, muhammet.demir\}@advocatehealth.org} \\
    {\tt\small \{Yongxin.Guo, Zhengjie.Zhu, Metin.Gurcan\}@wfusm.edu}
}

\maketitle

\thispagestyle{empty}
\pagestyle{empty}

\setcounter{figure}{0}\renewcommand{\thefigure}{S\arabic{figure}}
\setcounter{table}{0}\renewcommand{\thetable}{S\arabic{table}}
\setcounter{equation}{0}\renewcommand{\theequation}{S\arabic{equation}}

In this supplementary materials, we will include details
of datasets and implementation, as well as more experimental results.

\section*{S1. Datasets}
\label{sec:supp_datasets} 
We utilized two distinct datasets for this study: The Cancer Genome Atlas Breast Invasive Carcinoma (TCGA-BRCA) and an internal cohort.

\subsection*{TCGA-BRCA Dataset.}
Original Whole Slide Images (WSIs) in \texttt{.svs} format were downloaded from the official TCGA data portal. Our analysis was guided by the official clinical annotations for three tasks: Human Epidermal Growth Factor Receptor 2 (HER2) status, Progesterone Receptor (PR) status, and Oncotype DX (ODX) recurrence score. For the HER2 and PR tasks, we focused on the binary classification of positive versus negative status.

For the ODX risk stratification task, we used research-based ODX scores derived from normalized mRNA expression data, as calculated by~\cite{howardIntegrationClinicalFeatures2023}. The initial cohort of 1,133 WSIs underwent rigorous quality control, excluding cases with incomplete receptor status, missing ODX scores, gene profiles or processing failures. This resulted in a final analytical cohort of 997 WSIs. Within the hormone receptor-positive/HER2-negative (HR+/HER2-) subgroup ($n=516$), patients were categorized into low-risk ($n=443$) and high-risk ($n=73$) groups. The normalized ODX scores for this dataset range from $-2.009$ to $2.744$, with a risk threshold of $0.7169$. 

To ensure robust evaluation, patient-level stratification was employed to maintain a strict separation between training and testing cohorts, thereby preventing data leakage. To mitigate class imbalance issues, non-HR+/HER2- cases were also included in the training set. 

\subsection*{The In-House Dataset.}
This independent institutional cohort comprises 1,123 H\&E stained WSIs from HR+/HER2- breast cancer specimens. For this dataset, the ODX scores range from 0 to 100, with a clinical threshold of 25 used to differentiate low-risk ($n=961$) from high-risk ($n=162$) cases.

\begin{figure}[t]
\centering
\includegraphics[width=\linewidth]{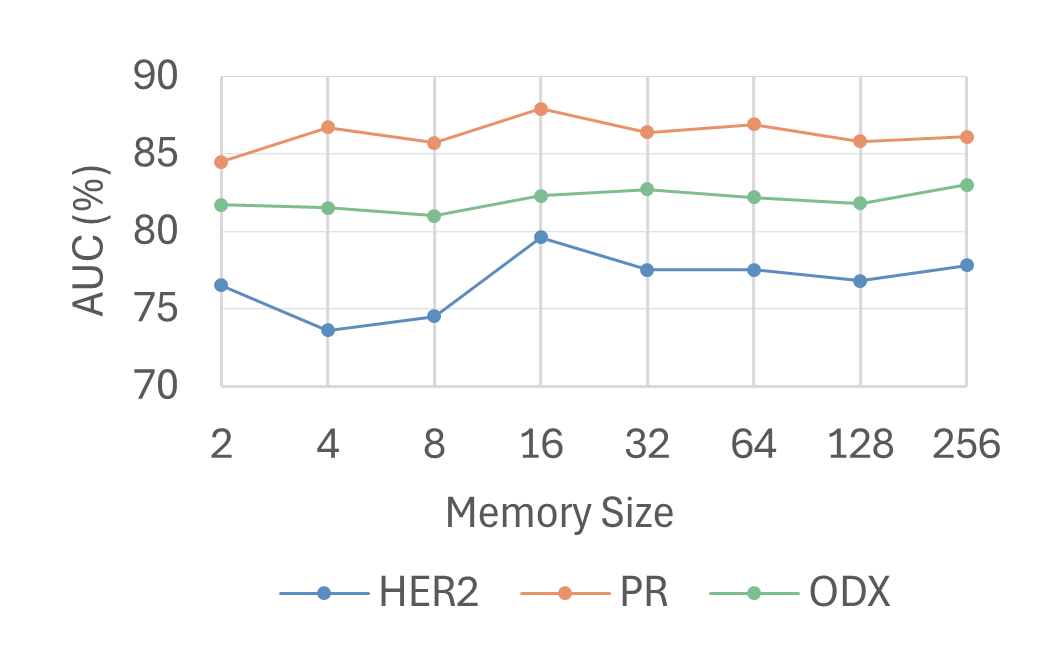}
\caption{The variation of memory size within three tasks.}
\label{fig:ksweep}
\end{figure}

\begin{figure*}[t]
  \centering
  \includegraphics[width=0.95\textwidth,keepaspectratio, trim=10 80 10 80, clip ]{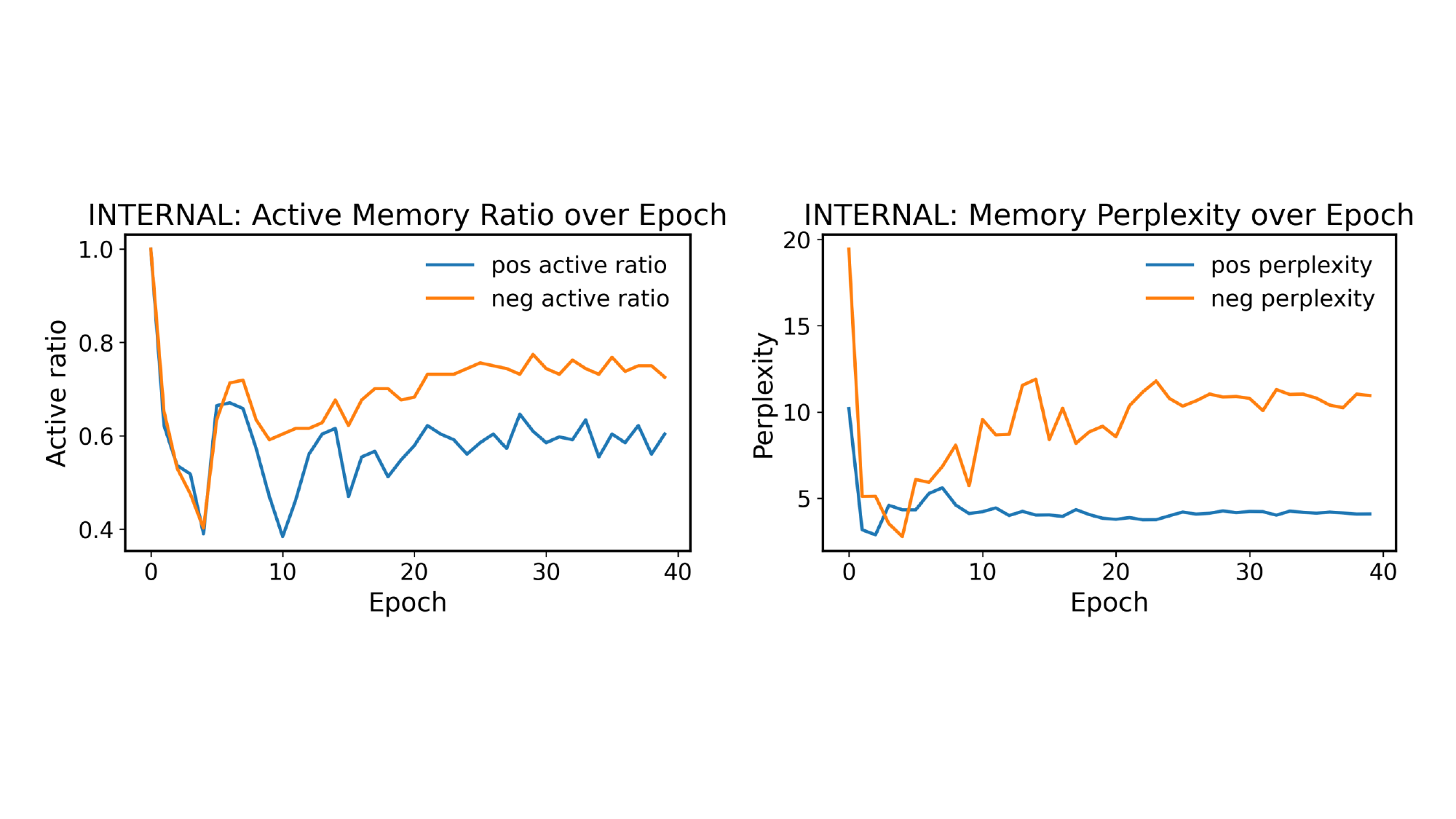}
  \caption{
 Memory dynamics on the in-house dataset.
  (Left) active memory ratio, and (right) perplexity evolution over training epochs.
  }
  \label{fig:internal_codebook}
\end{figure*}

\section*{S2. Implementation Details}
For WSI processing, we adopted the pipeline proposed by Trident~\cite{zhangAcceleratingDataProcessing2025}. Each WSI was initially segmented into non-overlapping tiles of $768 \times 768$ pixels. Features were then extracted from these tiles using the UNI v2 foundation model~\cite{chenGeneralpurposeFoundationModel2024}.

The original omics data presented a high-dimensional feature space with $D=19085$. To address this, we employed an XGBoost model for feature selection, identifying the top-$k$ genes, where $k$ was set to 768. The resulting data was subsequently normalized using the z-score method.

Our model was implemented in Python with the PyTorch library and trained on a single NVIDIA H-100 GPU. We utilized the AdamW optimizer with a learning rate of $2 \times 10^{-4}$ and a weight decay of $1 \times 10^{-4}$. The training was conducted for up to 80 epochs, incorporating an early stopping mechanism to select the best-performing checkpoint based on validation set accuracy. The batch size used is 1. For all tasks, the memory size was set to $n=16$. To stabilize training, model parameters were updated using a gradient accumulation strategy over 16 steps.

\section*{S3. Memory Dynamics on the Internal ODX Dataset}
In this section, we introduce more details on the memory analysis as well as its dynamics on the in-house dataset which shown in Fig.~\ref{fig:internal_codebook}.

\paragraph{Active Memory Ratio.}
\begin{equation}
r_{\mathrm{active}} = \frac{n_{\mathrm{active}}}{n},
\end{equation}
measures the proportion of memory mode entries updated within each epoch.
Values near 1.0 indicate broad participation, whereas smaller ratios reflect selective activation.
In our experiments, $r_{\mathrm{active}}>0.75$ throughout training, confirming the glocal activation.

\paragraph{Perplexity.}
\begin{equation}
\mathrm{Perplexity} = \exp(H(p)) \in [1, n],
\end{equation}
represents the effective number of active memory modes.
Perplexity values in different tasks reflect the dynamics learning process on the proposed method.

\paragraph{Clinical understanding and analysis}
The feasibility of inferring molecular biomarkers directly from H\&E stained whole slide image varies substantially across targets, reflecting the extent to which each biomarker manifests morphologically observable correlates. Among routinely assessed breast cancer biomarkers, HER2, progesterone receptor (PR), and Oncotype DX (ODX) recurrence scores represent three characteristic levels of morphological–molecular coupling.

HER2 (ERBB2 amplification and protein overexpression) is generally the most challenging target for image-based prediction~\cite{HER2TestingBreast,shamaiClinicalUtilityReceptor2024}. Its clinical definition relies on membranous protein overexpression and gene copy-number amplification assessed by immunohistochemistry (IHC) or in-situ hybridization (ISH), features that are not directly visible on standard H\&E slides~\cite{ivanovaStandardizedPathologyReport2024,valierisWeaklysupervisedDeepLearning2024}. Morphologic surrogates such as nuclear atypia, mitotic rate, or growth pattern provide only weak and indirect cues. Furthermore, even IHC-based HER2 scoring suffers from inter-observer variability , emphasizing its intrinsic diagnostic complexity.

In contrast, PR status tends to exhibit stronger alignment with histomorphological appearance. Hormone-receptor–positive tumors frequently present as low-grade, well-differentiated lesions with organized glandular architecture and lower mitotic activity that attributes readily captured in H\&E morphology~\cite{flanaganHistopathologicVariablesPredict2008}. 
\begin{algorithm*}[htbp]
\caption{Momentum Memory Knowledge Distillation (MoMKD)}
\label{alg:momkd_training}
\begin{algorithmic}[1]
\Require WSI spatial graph $G$, Omics vector $O$, Ground truth label $Y \in \{0, 1\}$
\Require Momentum memory banks $C^+$ (positive) and $C^-$ (negative)
\Statex
\State \textbf{// 1. Dual-Branch Modality Encoding}
\State $F_{wsi} \gets \text{WsiEncoder}(G)$ \Comment{Extract patch-level WSI representations}
\State $F_{omics} \gets \text{OmicsEncoder}(O)$ \Comment{Extract global omics representation}

\Statex
\State \textbf{// 2. Memory-Guided Distillation}
\State $Score \gets \text{ComputeAttention}(F_{wsi}, \text{Detach}(C^+), \text{Detach}(C^-))$ \Comment{Query memory for patch importance}
\State $F_{slide} \gets \text{Aggregate}(F_{wsi}, Score)$ \Comment{Obtain slide-level WSI representation}
\State $\hat{Y} \gets \text{Classifier}(F_{slide})$ \Comment{Generate diagnostic prediction}

\Statex
\State \textbf{// 3. Indirect Cross-Modal Alignment}
\State \textit{// Both modalities are aligned to the shared memory rather than to each other directly}
\State $L_{align}^{wsi} \gets \text{AlignmentLoss}(F_{wsi}, C^+, C^-, Y)$ \Comment{Align WSI to class-specific memory}
\State $L_{align}^{omics} \gets \text{AlignmentLoss}(F_{omics}, C^+, C^-, Y)$ \Comment{Align Omics to class-specific memory}

\Statex
\State \textbf{// 4. Omics Semantic Anchoring}
\State $L_{recon} \gets \text{ReconstructionLoss}(\text{Decoder}(F_{omics}), O)$ \Comment{Preserve biological structure}

\Statex
\State \textbf{// 5. Gradient-Decoupled Optimization}
\State $L_{task} \gets \text{CrossEntropy}(\hat{Y}, Y)$
\State 
\State \textit{// Decouple gradients to prevent modality collapse:}
\State \textbf{Update} $\text{WsiEncoder}$ and $\text{Classifier}$ using $\nabla (L_{task} + L_{align}^{wsi})$
\State \textbf{Update} $\text{OmicsEncoder}$ using $\nabla (L_{recon} + L_{align}^{omics})$ 
\State \textbf{Update} $C^+, C^-$ using $\nabla (L_{align}^{wsi} + L_{align}^{omics} + L_{mem})$ \Comment{Shielded from $L_{task}$}
\end{algorithmic}
\end{algorithm*}

Finally, the Oncotype DX recurrence score, a multigene assay quantifying proliferation and differentiation-related transcripts which shows the closest association with H\&E-derived phenotypes. Multiple clinical studies have demonstrated strong correlations between ODX and conventional histologic variables such as tumor grade, nuclear pleomorphism, and mitotic index~\cite{whitneyQuantitativeNuclearHistomorphometry2018a,suComputationalPathologyAccurate2024c,paikMultigeneAssayPredict2004}. Deep learning approaches leveraging WSIs and minimal clinical data have achieved concordance comparable to molecular testing (AUC $\approx 0.80–0.85$ for high- vs low-risk classification~\cite{guoBPMambaMILBioinspiredPrototypeguided2025,suBCRNetDeepLearning2023}). These findings suggest that ODX captures transcriptomic programs that are largely mirrored by morphological cues observable in H\&E slides.

In summary, the variable predictability of HER2, PR, and ODX from H\&E images arises from the differing degrees of morphological expressivity of their underlying biology. HER2 overexpression reflects membrane-localized molecular events poorly represented in tissue architecture; PR status influences global differentiation that leaves discernible patterns; and ODX aggregates proliferation-related signals that are morphologically pronounced. Acknowledging this gradient of morphologic–molecular coupling is essential when interpreting model performance.

Based on the statistics in Fig.~\ref{fig:internal_codebook} and the clinical characteristics of each biomarker, here we provide additional insight into the development of the memory used in our framework. The memory usage both in the TCGA-BRCA dataset (Figure~3) with the in-house dataset (Figure~S2) indicate the memory component activation remains consistently broad, with over 75\% of memory components active across epochs. This directly validates the efficacy of the gradient-decoupled momentum update, proving that the memory maintains rich semantic diversity and successfully avoids global collapse that often plaguing dynamic dictionaries.

Interestingly, task-dependent memory usage patterns emerge. The HER2 classification task activates a greater variety of memory usage, consistent with its higher histopathology complexity and known diagnostic ambiguity on H\&E stains~\cite{shamaiClinicalUtilityReceptor2024}. In contrast, PR and ODX tasks exhibit more concentrated usage, suggesting that fewer memory components suffice to capture discriminative features in these more visually distinguishable tasks~\cite{suComputationalPathologyAccurate2024b}. This behavior aligns with the notion of functional sparsity: the model automatically compresses its memory usage when the task allows, while expanding representational breadth for histopathology heterogeneous conditions. These findings highlight that the momentum memory offering a dynamic balance between expressivity and compactness that is absent in static-memory or batch-local alignment methods.

\begin{figure}[t]
  \centering
  \includegraphics[width=0.95\columnwidth,keepaspectratio, trim=180 180 180 110, clip ]{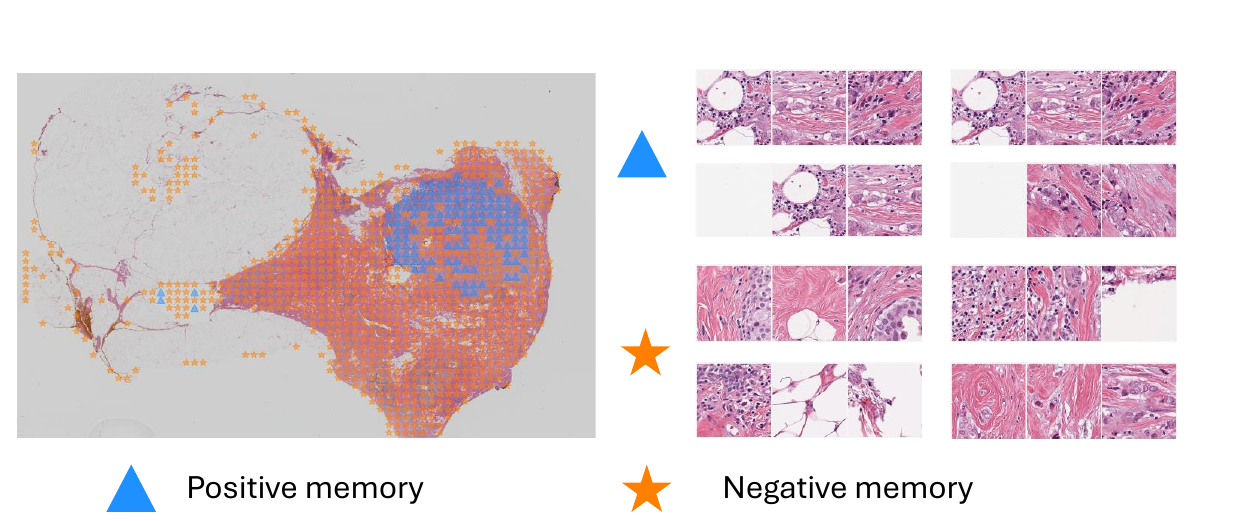}
  \caption{
  The visualization of memory usage on the misclassified case from the TCGA-BRCA dataset.
  }
  \label{fig:false_case}
\end{figure}

\section*{S4. Visualization on the misclassified case}
To further investigate the misclassifications, we visualize the memory components of failure cases from the TCGA-BRCA dataset in the HER2 task (Fig.~\ref{fig:false_case}). As illustrated, the positive memory disproportionately attends to patches dominated by non-informative white background—redundant regions that theoretically should have been eliminated during initial preprocessing. This erroneous focus is similarly reflected in the negative memory representations. Consequently, these results highlight that rigorous background filtering and precise feature extraction remain crucial bottlenecks in standard WSI processing pipelines.

\section*{S5. Pseudo code for the proposed method}
In the Algorithm 1, we provide the pseudo code for the MoMKD. The detailed implementation can be found in our GitHub repo.

{
   \small
   \bibliographystyle{ieeenat_fullname}
   \bibliography{library}
}